\pdfoutput=1

\documentclass[11pt]{article}

\PassOptionsToPackage{dvipsnames,table,x11names}{xcolor}
\usepackage{xcolor}

\usepackage[final]{acl}

\usepackage[T1]{fontenc}        
\usepackage[utf8]{inputenc}     
\usepackage{times}              
\usepackage{latexsym}           
\usepackage{inconsolata}        
\usepackage[cjk]{kotex}         

\usepackage{graphicx}           

\usepackage{amsmath}            
\usepackage{amssymb}            
\usepackage{mathtools}          
\usepackage{bm}                 

\usepackage{booktabs}           
\usepackage{multirow}           
\usepackage{hhline}             
\usepackage{tabularx}           
\usepackage{colortbl}           

\usepackage{subcaption}         

\usepackage{algorithm}          
\usepackage{algpseudocode}      

\usepackage{enumitem}           

\usepackage{setspace}           
\usepackage{xspace}             
\usepackage[normalem]{ulem}     
\usepackage{layouts}            
\usepackage{footnote}
\usepackage[verbose=false]{microtype}

\usepackage{pifont}             

\usepackage[most]{tcolorbox}    

\usepackage[%
    commandnameprefix=always,   
    defaultcolor=purple         
]{changes}

\usepackage[outline]{contour}   

\useunder{\uline}{\ul}{}
\definecolor{maincolor}{RGB}{171, 225, 161}
\definecolor{orangebg}{RGB}{255, 244, 230}
\definecolor{orangetxt}{RGB}{229,90,0}
\definecolor{bluebg}{RGB}{230,241,255} 
\definecolor{bluetxt}{RGB}{43,85,157}
\definecolor{promptgray}{RGB}{250, 250, 250}

\definecolor{zoominred}{RGB}{255, 217, 206}
\definecolor{zoomoutblue}{RGB}{215, 231, 253}
\definecolor{originyellow}{RGB}{252, 242, 200}

\definecolor{bluetxt}{RGB}{43,85,157}

\newcommand{\aspect}[1]{\colorbox{orangebg}{#1}}
\newcommand{\opinion}[1]{\colorbox{bluebg}{#1}}

\newcommand{\inbox}[1]{\colorbox{zoominred}{#1}}
\newcommand{\outbox}[1]{\colorbox{zoomoutblue}{#1}}
\newcommand{\originbox}[1]{\colorbox{originyellow}{#1}}

\newcommand{\judge}{\textsc{Judge}\xspace}
\newcommand{\narrow}{\textsc{Zoom-In}\xspace}
\newcommand{\widen}{\textsc{Zoom-Out}\xspace}
\newcommand{\method}{\textsc{Zoom In-N-Out}\xspace}

\newcommand{\eg}{\emph{e}.\emph{g}.,\xspace}
\newcommand{\ie}{\emph{i}.\emph{e}.,\xspace}

\title{
Single Ground Truth Is Not Enough: \\
Adding Flexibility to Aspect-Based Sentiment Analysis Evaluation
}

\author{\textbf{Soyoung Yang}$^\heartsuit$ \hspace{0.3cm}
        \textbf{Hojun Cho}$^\heartsuit$ \hspace{0.3cm}
        \textbf{Jiyoung Lee}$^\heartsuit$ \hspace{0.3cm}
        \textbf{Sohee Yoon}$^\clubsuit$ \hspace{0.3cm} \\
        \textbf{Edward Choi}$^\heartsuit$ \hspace{0.3cm}
        \textbf{Jaegul Choo}$^\heartsuit$ \hspace{0.3cm}
        \textbf{Won Ik Cho}$^\clubsuit$ \hspace{0.3cm} \\[0.1cm]
        $^\heartsuit$ KAIST AI \hspace{0.3cm}
        $^\clubsuit$ AI Center, Samsung Electronics \\
        \texttt{sy\_yang@kaist.ac.kr} \\
}

\begin{document}
\maketitle

\def\thefootnote{$^\clubsuit$}\footnotetext{Work is done while affiliated to SAIT.}\def\thefootnote{\arabic{footnote}}
\thispagestyle{empty}

\begin{abstract}

Aspect-based sentiment analysis (ABSA) is a challenging task of extracting sentiments along with their corresponding aspects and opinion terms from the text.
The inherent subjectivity of span annotation makes variability in the surface forms of extracted terms, complicating the evaluation process.
Traditional evaluation methods often constrain ground truths (GT) to a single term, 
potentially misrepresenting the accuracy of semantically valid predictions that differ in surface form.
To address this limitation, we propose a novel and fully automated pipeline that expands existing evaluation sets by adding alternative valid terms for aspect and opinion. 
Our approach facilitates an equitable assessment of language models by accommodating multiple-answer candidates, resulting in enhanced human agreement compared to single-answer test sets (achieving up to a 10\%p improvement in Kendall's Tau score).
Experimental results demonstrate that our expanded evaluation set helps uncover the capabilities of large language models (LLMs) in ABSA tasks, which is concealed by the single-answer GT sets.
Consequently, our work contributes to the development of a flexible evaluation framework for ABSA by embracing diverse surface forms to span extraction tasks in a cost-effective and reproducible manner.
Our code and dataset is open at \url{https://github.com/dudrrm/zoom-in-n-out-absa}.
\end{abstract}

\section{Introduction}

\begin{figure}[t!]
\begin{center}
\includegraphics[width=\columnwidth]{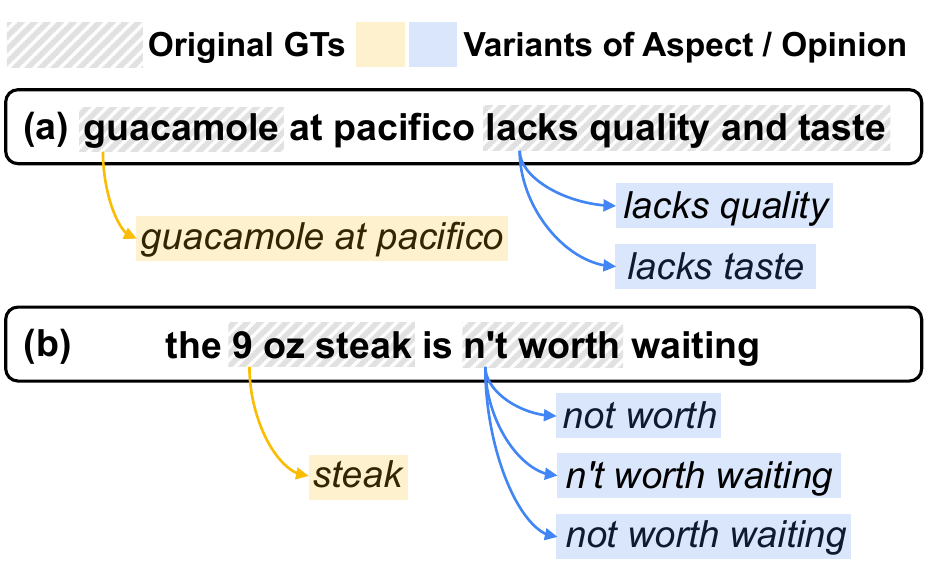}
\end{center}
\vspace{-2mm}
\caption{
Original ground truth (GT) aspect and opinion terms are shown in gray background shading. 
Alternative valid expressions for these terms are highlighted: aspects in yellow and opinions in blue.
Under conventional evaluation approaches using the original GT set, these highlighted candidates are assessed as \textit{wrong} because they do not match exactly despite being semantically consistent with the original terms.
}
\label{fig:teaser-observation}
\end{figure}

Aspect-based sentiment analysis (ABSA) is a sophisticated natural language processing task that aims to extract fine-grained sentiment information from text.
ABSA is usually utilized to provide insights into opinions about specific attributes of products or services, enabling organizations or persons to precisely identify consumer preferences and criticisms. 
The granularity of ABSA facilitates targeted improvements and data-driven decision-making across diverse domains, including product development, customer service, political analysis, and e-commerce~\cite{ha-2019-absa-dl-review,zhang2022survey}.
ABSA includes a subtask known as aspect sentiment quadruple prediction (ASQP), which involves identifying four elements within the text: (aspect, category, sentiment, opinion).
For illustration, consider the sentence in Figure~\ref{fig:teaser-observation}~(a).
From this, we can extract the quadruple of (``guacamole'', Food Quality, Negative, ``lacks quality and taste'').
Here, the gray-shaded terms represent the ground truth (GT) annotations: ``guacamole'' as the aspect and ``lacks quality and taste'' as the opinion.

Due to the richness and variability of natural language, there are diverse forms in which these elements are expressed and consequently annotated.
Different annotators may select different spans to represent the same semantic content, producing various surface forms in span annotations.
In Figure~\ref{fig:teaser-observation}, the terms highlighted in color illustrate alternative candidates that can be considered valid annotations.

However, conventional ABSA benchmarks~\cite{pontiki-etal-2014-semeval,pontiki-etal-2015-semeval, pontiki-etal-2016-semeval,zhang-2021-asqp-paraphrase,cai-2021-acos} typically provide a single GT term for each aspect and opinion pair and employ exact match criteria for evaluation.
This approach fails to account for the compatibility of multiple surface forms.
As illustrated in Figure~\ref{fig:teaser-observation}, semantically valid surface forms (highlighted in color), such as ``guacamole at pacifico'' for the aspect or ``lacks quality'' and ``lacks taste'' for the opinion, are evaluated as \textit{incorrect}, while they preserve the essential meaning of the original GT quadruple.
Consequently, this rigid evaluation schema may underestimate the performance of advanced language models in understanding and generating semantically valid predictions, raising concerns about the fairness and comprehensiveness of current model assessments in ABSA.

To address the limitations of conventional ABSA evaluation approaches, which often fail to account for the variability of natural language expressions, we propose \method, a novel approach for evaluation in ABSA task~\footnote{While implicit terms (\eg``null'') are considered in ABSA benchmarks~\cite{zhang-2021-asqp-paraphrase,cai-2021-acos}, our method aims to expand explicit GT expressions in aspect and opinion terms. We leave this exploration for future work.}.
Our method aims to accommodate diverse surface forms that preserve the semantic content of the original GT quadruple.
\method pipeline comprises three key steps: (i) identifying alternative terms within the original GT span, (ii) exploring potential expressions outside the span, and (iii) verifying that new candidates are semantically consistent with the original GT.
This process leverages the capabilities of large language models (LLMs), specifically GPT-4, which has demonstrated quality in annotation tasks~\cite{Gilardi2023gptoutperformcrowdworkers} and efficacy in generating synthetic datasets~\cite{zheng2023MTBench-neurips,alpaca_eval}.
Notably, our approach offers several advantages over existing methods, such as N-gram overlap metrics or direct LLM/human evaluation. 
N-gram-based methods like F-measure~\cite{melamed-etal-2003-precision-f-score} may fail to capture semantic nuances; for instance, they may incorrectly score ``not worth'' and ``worth'' as similar when GT is ``n't worth''. 
Direct LLM/human evaluation~\cite{zheng2023MTBench-neurips,wadhwa-etal-2023-revisiting} can be costly and difficult to reproduce since the evaluator model should be run when a new prediction occurs. 
In contrast, our method provides a balanced solution by ensuring semantic consistency with the original GT and employing LLMs solely during the GT expansion phase, allowing the subsequent evaluations to be cost-effective and reproducible.

Experimental results demonstrate the quality and effectiveness of our method.
Our expanded GT set achieves over 90\% validity in human assessment and shows improved human alignment, with up to a 10\%p increase in Kendall Tau score compared to the original GT set. 
Furthermore, compared to T5-based models, the F1 scores of LLMs jump up in our evaluation set when the train set (in-context examples) comes from the original dataset.
For instance, four LLMs increase their F1 score by 9.8\%p, but T5 models increase by only 2.3\%p on average.
This significant score gap suggests that the conventional evaluation framework may have underestimated the capabilities of LLMs.
It highlights how our method uncovers their hidden ability through our flexible evaluation framework.

In summary, the contributions of our paper can be summarized as follows:

\begin{itemize}[topsep=-0.5pt, itemsep=0ex, partopsep=.8ex, parsep=.8ex]

    \item We introduce \method, a fully automated pipeline that addresses the diversity of expressions in the ABSA ground truth set by expanding aspect and opinion terms in existing datasets.
    
    \item Our experimental results demonstrate that our expanded truth set aligns closely with human judgments more than the conventional test set, highlighting that our dataset has effectively reflected the linguistic variability in ABSA evaluation.
    
    \item We conduct an extensive comparative study of ABSA performance between LLMs and ABSA-specific T5 models, revealing the underestimated potential of LLMs in ABSA tasks and proposing a more equitable evaluation framework for the LLM era.
    
    \end{itemize}

\begin{figure*}[!]
\begin{center}
\includegraphics[width=\textwidth]{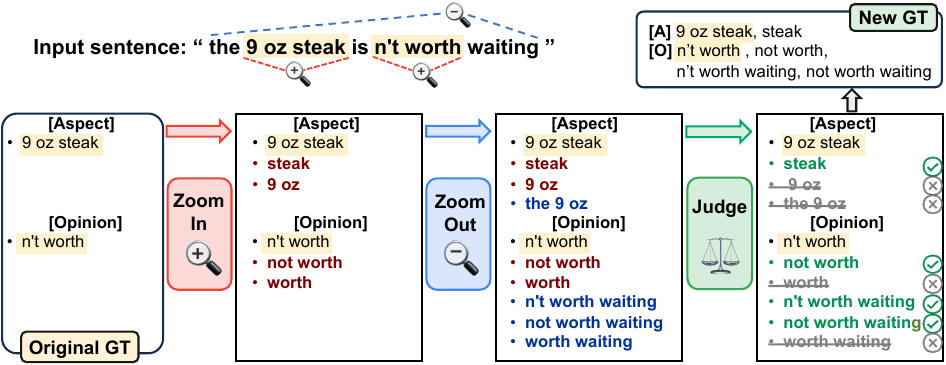}
\end{center}
\caption{
Overview of \method, illustrating the extraction of diverse candidates of the aspect and opinion terms in ABSA task.
The three stages of our pipeline are depicted by three colored blocks.
Starting with the input sentence and the original ground truth term highlighted in yellow, the model generates alternative representations of the target term at each stage. 
The original ground truth quadruple is (``9 oz steak'', Food Quality, Negative, ``n't worth''), representing aspect, category, sentiment, and opinion, respectively.
Since our pipeline preserves the provided categorical values, the category and sentiment are not included in the figure.
}
\label{fig:pipeline}
\end{figure*}

\section{Related work}

\subsection{LLM in Information Extraction}

The advent of proprietary LLMs, such as GPT~\cite{NEURIPS2020GPT3-llmsarefewshotlearners} and Gemini~\cite{geminiteam2024gemini}, along with open-sourced models like Llama~\cite{touvron2023llama,dubey2024llama3.1} and its variants~\cite{alpaca,vicuna2023}, has led to their widespread adoption in natural language processing (NLP)~\cite{liu-etal-2022-p,min-etal-2022-rethinking,yoo-etal-2022-labelsdoesmatter-ICL}. 
In information extraction tasks, LLMs have been applied to various tasks, including named entity recognition~\cite{xie2023selfimproving-ner,wu2024consistentNER}, relation extraction~\cite{wan2023gpt-re,wadhwa-etal-2023-revisiting,ma-etal-2024-star-re}, and event extraction~\cite{pang-etal-2023-guideline-ee,wang-etal-2023-code4struct-ee}. 
Data augmentation with LLMs has also been actively studied~\cite{xu-etal-2023-unleash-da-re,wang-etal-2023-improving-unsupervised-aug-re}. 
While prevalent data augmentation approaches~\cite{zhang-et-al-2015-characterCNN-DA,kobayashi-2018-contextual-DA} aim to improve training performance by increasing the training set size, our work focuses on re-evaluating model performance by considering diverse surface forms in the evaluation set.

\begin{table}[t]
\centering
\resizebox{\columnwidth}{!}{%
\begin{tabular}{ll}
\toprule 
\textbf{Task} & \textbf{Output} \\ \midrule 
Aspect term extraction (ATE) & $a$ \\
Aspect-opinion pair extraction (AOPE) & $a, o$ \\
Aspect-sentiment pair extraction (ASPE) & $a$, {$s$} \\
Aspect-sentiment triplet extraction (ASTE) & $a, o,$ $s$ \\
Aspect-category-sentiment detection (ACSD) & $a$, $c$ $s$ \\
\textbf{Aspect-sentiment quad prediction (ASQP)} & $a$, $c$, $s$, $o$
\\ \bottomrule
\end{tabular}%
}
\caption{Target output elements for ABSA tasks. In ASQP, the output is represented as a quadruple: (aspect term $a$, aspect category $c$, sentiment polarity $s$, opinion term $o$). The names of the tasks follow \citet{zhang2022survey}.}
\label{tab:absa-subtasks-def}
\end{table}
\subsection{Aspect-based Sentiment Analysis}

Current ABSA datasets are primarily derived from SemEval tasks, including aspect-sentiment triplet extraction~\cite{pontiki-etal-2014-semeval,pontiki-etal-2015-semeval,pontiki-etal-2016-semeval} and target-aspect sentiment joint detection~\cite{saias-2015-sentiue}. 
As shown in Table~\ref{tab:absa-subtasks-def}, ABSA started as an aspect-based sentiment analysis but has evolved into different formats, such as target-oriented opinion words extraction~\cite{fan-etal-2019-AO-towe-dataset}, aspect-sentiment quad prediction (ASQP)~\cite{zhang-2021-asqp-paraphrase,cai-2021-acos}.
In this paper, we focus on the ASQP task, where the output is a set of quadruples.

With the advent of pre-trained language models, BERT-based architectures~\cite{devlin-etal-2019-bert} have become dominant in ABSA research~\cite{zhang2022survey,cai-2023-memdabsa,gao-etal-2022-lego-absa,wang-2024-unifiedABSA}.
Recent ASQP approaches have adopted T5-style encoder-decoder architectures~\cite{raffel2020exploring}, including Paraphrase~\cite{zhang-2021-asqp-paraphrase}, which derives sentence structures fitting quadruple prediction; DLO~\cite{hu-etal-2022-dlo-template}, which augments datasets through element permutation; and MvP~\cite{gou-etal-2023-mvp}, which creates and ensembles quadruples of multiple element orders for majority-voted predictions. 
Including DLO and MvP, various studies~\cite{li-etal-2020-conditional-DA,hsu-etal-2021-semantics-DA} utilize data augmentation approaches to alter the training set and improve model performance.
Our approach, however, differs significantly. 
We focus on establishing a fair ABSA evaluation framework by expanding the evaluation set to account for linguistic variability in aspect and opinion terms, without modifying the original input sentences or training dataset.

\section{Method}

Our pipeline aims to expand the original GT aspect and opinion terms to various equivalent expressions, to encompass the diversity of surface forms in the ABSA task. 
Based on Figure~\ref{fig:pipeline}, we introduce \method and describe how the exact match measure is calculated on our GT sets.

\subsection{\method}
Let $D = \{(x_k, Y_k)\}_{k=1}^{N_\text{data}}$ denote test set, where $N_{\text{data}}$ represents the total number of test examples.
Consider $d = (x, Y) \in D$ where $Y$ comprises one or more ground truth (GT) quadruples, such that $Y = \{y_i\}_{i=1}^{N}$ for $N \geq 1$.
Each GT quadruple $y \in Y$, derived from $x$, is defined as $(a, c, s, o)$, representing aspect, category, sentiment, and opinion, respectively.
As illustrated in Figure~\ref{fig:pipeline}, our pipeline consists of three steps separately applied to aspect and opinion terms before merging the results.
In the following section, we will explain each step, focusing on the case of the opinion term, \ie ``n't worth'' from Figure~\ref{fig:pipeline}.

\paragraph{\narrow}
Beginning with the original GT opinion term $o$ in $y$ and the given sentence $x$, the LLM (hereafter `the model') investigates the inside of $o$, either by reshaping the given term or extracting meaningful components.
From a given opinion term, the model processes ``n't worth'' to generate (i) ``not worth'' by resolving the contraction ``n't'' to ``not'' and (ii) ``worth'' as a partial component.

\paragraph{\widen}
Following the internal examination of $o$, we guide the model to explore the entire sentence $x$ to identify and incorporate adjacent words.
From a given sentence, the model extracts combinations such as ``n't worth waiting'', ``not worth waiting'', and ``worth waiting'' by integrating ``waiting'', which follows the original opinion term.

\paragraph{\judge and Filter}
Newly generated terms may not always be appropriate due to the hallucination.
To exclude unsuitable terms, we establish four criteria:
(i) relevance to aspect and category, (ii) consistency with opinion and sentiment, (iii) extractability from $x$, and (iv) independence from other GT terms, which is based on the SemEval annotation guideline~\cite{pontiki-etal-2016-semeval}.
Following the LLM-as-a-judge~\cite{zheng2023MTBench-neurips,madaan2023selfrefine}, the model verifies if a new term meets all four criteria.
We exclude the new term if any of the criteria are unmet.
In Figure~\ref{fig:pipeline}, ``worth'' and ``worth waiting'' are excluded for violating the second criterion by failing to maintain the negative sentiment of the original term.

After completing the same procedure for both aspect and opinion terms, we obtain an aspect term set $A = \{a_n\}_{n=1}^{N_a}$ and an opinion term set $O=\{o_m\}_{m=1}^{M_o}$, where $N_a$ and $M_o$ represent the number of aspect and opinion terms, respectively. 
Each set includes the original term as its first. 
The new GT ($y^{\text{new}}$) becomes a list of $(a_n, c, s, o_m)$ where $a_n \in A$ and $o_m \in O$, yielding $N_a \times M_o$ combinations for each aspect and opinion term.
By applying the pipeline to all  $y_i \in Y$, we obtain $Y^{\text{new}} = \{y_i^{\text{new}}\}_{i=1}^{N}$ where each $y_i^{\text{new}}$ contains all new combinations for $y_i$.
We perform simple post-processing to remove potential duplicates within $Y^{\text{new}}$.
Repeating this procedure across all $d_k =(x_k,Y_k) \in D$, we achieve an expanded test set $D^{\text{new}} = \{(x_k, Y_k^{\text{new}})\}_{k=1}^{N_\text{data}}$.

Algorithm~\ref{alg:method} formalizes our process, with newly added components denoted by ($'$).
All three functions \narrow, \widen, and \judge correspond to the above steps, taking $E \in \{A, O\}$ as input and updating it based on the given $x,y$ and the established criteria.
While diverse prompting methods can be applied at each step of \narrow, \widen, and \judge, we adopt a straightforward approach: 
5-shot ICL for \narrow and \widen steps and 5-shot chain-of-thought (CoT) for \judge. 
The specific details of our experimental design are elaborated in Section~\ref{sec:experiments}.
Detailed prompts and examples are provided in Appendix~\ref{sec:Prompts}.

\begin{algorithm}[t]
  \caption{\method} 
  \label{alg:method}
  \small
  \begin{algorithmic}[1]
    \State \textbf{Input:} $D = \{d_i\} = \{(x_k, Y_k)\}$
    \For{$d = (x, Y = \{y\}) \in D$}
      \State $Y' \gets []$
      \For{$y = (a,c,s,o) \in Y$} 
        \State $A \gets \{a\}$, $O \gets \{o\}$
        \For{$E \in \{A, O\}$}
            \State $E \gets \Call{Zoom-In}{x, y, E}$
            \State $E \gets \Call{Zoom-Out}{x, y, E}$
            \State $E \gets \Call{Judge}{x, y, E}$  
        \EndFor
        \State $y' \gets \{(a',c,s,o') \mid a' \in A, o' \in O\}$
        \State $y' \gets \text{Filter}(y')$ 
        \State $Y' \gets Y' + y'$
      \EndFor
      \State $d' \gets (x, Y')$
    \EndFor
    \State \textbf{Output:} $D^\text{new} = \{d'_i\} = \{(x_k, Y'_k)\}$
  \end{algorithmic}
\end{algorithm}

\subsection{Measurement}
For a given task $T$, here ABSA (focusing on the ASQP task),
let a sentence $x$ be an input sentence for a language model $f$ that predicts one or more quadruples.
Given the principles of $T$, each $x$ has a ground truth ($Y$), a set of quadruples {$y_i$} where each $y_i$ consists of $(a, c, s, o)$.
Let $\hat{Y} = f(x)$ be the prediction, a set of predicted quadruples \{$\hat{y}_j$\}.
$\hat{y}_j$ is considered a true positive only if $\hat{y}_j$ in $Y$, following the conventional exact match ($EM$) criterion.

Given our process \method as $Z$ applied to $D$ such that $D^\text{new} = Z(D)$, we calculate $F1(f,D^\text{new};T,EM)$ instead of the original $F1(f,D;T,EM)$.
Our method employs the conventional metrics, \ie $EM$ and $F1$, in the same manner,
with the key difference being the expansion of each ground truth $Y = {y_i}$ to $Y^{\text{new}} = {y_i^\text{new}}$, where $y_i \in y_i^{\text{new}}$ for every $y_i$ and $y_i^\text{new}$.
This expansion does not necessarily increase the number of effective GTs, but rather broadens the scope of quadruple candidates counted as a GT.

\section{Experiments}
\label{sec:experiments}
In this section, we first assess the validity of our expanded GTs, show that our evaluation set aligns more closely with human evaluators than the original, and analyze diverse baseline models with the original and our GT sets.

\subsection{Experimental Setting}

\paragraph{Construction of Our Evaluation Set}
We expand the GT quadruples of the ACOS and ASQP test sets with \method.
The pipeline processes each aspect and opinion term independently before merging the results.
All the expansion processes utilize \textit{gpt-4o-2024-05-1}, with an average cost of \$75 per extended dataset.
\narrow and \widen steps employ 5-shot ICL to collect possible quadruples (examples in Table~\ref{tab:icl_prompt}).
To maximize the diversity, we perform three generations with a temperature of 0.3.
\judge uses 5-shot CoT prompting to verify each quadruple (examples in Table~\ref{tab:judge_ouptut}).

\paragraph{Datasets}
ASQP~\cite{zhang-2021-asqp-paraphrase} and ACOS~\cite{cai-2021-acos} datasets are utilized.
For ACOS-Laptop, which contains 121 total categories, we focus on the 23 entity-level categories, since the category comprises two levels: entity, \eg ``laptop'', and attribute, \eg ``price''.

\paragraph{Base models}
We evaluate both T5 and LLM-based ABSA models in a few-shot setting to compare the original GT set $Y$ with our extended set $Y^{\text{new}}$ produced by \method.
All experiments use the same 20 examples per dataset, and the heuristic selection criteria can be found in Appendix~\ref{sec:append-demo}.
Representative T5-based fine-tuning methodologies for ABSA include Paraphase~\cite{zhang-2021-asqp-paraphrase}, DLO~\cite{hu-etal-2022-dlo-template}, and MVP~\cite{gou-etal-2023-mvp}.
Following the low-resource learning protocol of MvP, each model undergoes initial fine-tuning on the ASTE dataset~\cite{peng2020-aste-dataset} for 20 epochs, followed by transfer learning with 20 shots on each dataset for 100 epochs.
Each model is trained five times with different seeds to measure the average performance.
For LLMs, we employ several LLMs with 20-shot ICL: GPT-3.5-Turbo in \textit{gpt-3.5-turbo-0125}, GPT-4o-mini in \textit{gpt-4o-mini-2024-07-18}, Gemini-v1.5-Pro~\cite{geminiteam2024gemini}, and Llama-3.1-70B-Instruct~\cite{dubey2024llama3.1}~\footnote{The experiments are done in April, May, September, and October 2024.}.
System prompts are derived from the SemEval annotation guidelines~\cite{pontiki-etal-2016-semeval}.
LLM-based predictions utilize greedy decoding with temperature set to 0.
The few-shot examples are provided in Table~\ref{tab:demonstration_examples}.

\begin{table}[t]
\centering
\resizebox{.9\columnwidth}{!}{%
\begin{tabular}{l|cc|cc}
\toprule
 & \multicolumn{2}{c|}{\textbf{\# Preds.}} & \multicolumn{2}{c}{\textbf{\# GTs.}} \\
\textbf{Datasets} & \textbf{MvP} &  \textbf{GPT 3.5} & \textbf{Orig.} & \textbf{New} \\
 \midrule
\textbf{ASQP-R15} & 137 & 147 & 128 & 418 \\
\textbf{ASQP-R16} & 120 &  141 & 110 & 292 \\
\textbf{ACOS-Laptop} & 108 &  131 & 114 & 255 \\
\textbf{ACOS-Rest} & 137 & 15 & 132 & 355
\\ \bottomrule
\end{tabular}%
}
\caption{Number of quadruples utilized in human evaluation study.
The predicted quadruples for two models (\#Preds.) and the number of evaluated GT quadruples (\#GTs).
}
\label{tab:user-study-n-p}
\end{table}

\begin{table}[t]
\resizebox{\columnwidth}{!}{%
\begin{tabular}{l|cccc}
\toprule
 & \textbf{ASQP-R15} & \textbf{ASQP-R16} & \textbf{ACOS-L} & \textbf{ACOS-R} \\
 \midrule
\textbf{Agr}. & 99.04 & 97.95 & 96.47 & 93.52
\\ \bottomrule
\end{tabular}%
}
\caption{Human agreement (\%) on the validity of our expanded test set. ACOS-L and ACOS-R represent ACOS-Laptop and ACOS-Rest, respectively.}
\label{tab:validity-agreement}
\end{table}

\begin{table*}[ht]
\small
\resizebox{\textwidth}{!}{%
\begin{tabular}{c||cc|cc|cc|cc}
\toprule
\multirow{2}{*}{\textbf{Models}} & \multicolumn{2}{c|}{\textbf{ASQP-Rest15}} & \multicolumn{2}{c|}{\textbf{ASQP-Rest16}} & \multicolumn{2}{c|}{\textbf{ACOS-Laptop}} & \multicolumn{2}{c}{\textbf{ACOS-Rest}} \\
 & \textbf{Orig.} & \textbf{Ours} & \textbf{Orig.} & \textbf{Ours} & \textbf{Orig.} & \textbf{Ours} & \textbf{Orig.} & \textbf{Ours} \\
\midrule
MVP & 33.4 / 42.3 & \textbf{39.5 / 47.4} & 56.6 / 61.8 & \textbf{58 / 64.1} & 47.6  /  58.1 & \textbf{52.7 / 59.8} & 47.3 / 54.5 & 47.3 / 54.5 \\
GPT-3.5-Turbo  & 8.3 / 18.8 & \textbf{18 / 29.8} & 26.3 / 38.9 & \textbf{48.9 / 48.9} & 12.8 / 22.5 & \textbf{15 / 23.7} & 23.5 / 33.9 & \textbf{31.1 / 40.6} \\
\bottomrule
\end{tabular}%
}
\caption{
Cohen's Kappa and Kendall tau ($\kappa$ / $\tau$) between the human evaluators and the two GT sets when assessing the models' predictions.
Scores are multiplied by 100.
}
\label{tab:human-agreement}
\end{table*}

\begin{table*}[t]
\centering
\small
\resizebox{\textwidth}{!}{%
\begin{tabular}{l|ccc|ccc|ccc|ccc}
\toprule
\multicolumn{1}{c|}{} & \multicolumn{3}{c|}{\textbf{ASQP-Rest15}} & \multicolumn{3}{c|}{\textbf{ASQP-Rest16}} & \multicolumn{3}{c|}{\textbf{ACOS-Laptop}} & \multicolumn{3}{c}{\textbf{ACOS-Rest}} \\
\multicolumn{1}{c|}{\multirow{-2}{*}{\textbf{Models}}} & \textbf{Orig.} & \textbf{Ours} & \textbf{$\Delta$} & \textbf{Orig.} & \textbf{Ours} & \textbf{$\Delta$} & \textbf{Orig.} & \textbf{Ours} & \textbf{$\Delta$} & \textbf{Orig.} & \textbf{Ours} & \textbf{$\Delta$} \\
\\[-2ex]
\hline
\rowcolor[HTML]{EFEFEF}
\multicolumn{13}{l}{\cellcolor[HTML]{EFEFEF}\textit{\textbf{Naive}}} \\

Paraphrase & 22.55 & 25.93 & \cellcolor{maincolor!7}3.38 & 31.27 & 33.62 & \cellcolor{maincolor!4}2.35 & 8.97 & 9.53 & \cellcolor{maincolor!0}0.56 & 22.91 & 24.17 & \cellcolor{maincolor!0}1.26 \\
DLO & 31.17 & 35.7 & \cellcolor{maincolor!15}4.53 & 37.84 & 42.52 & \cellcolor{maincolor!22}4.68 & 12.44 & 13.32 & \cellcolor{maincolor!6}0.88 & 34.3 & 35.82 & \cellcolor{maincolor!2}1.52 \\
MvP & 33.95 & 36.97 & \cellcolor{maincolor!4}3.02 & 42.26 & 45.05 & \cellcolor{maincolor!8}2.79 & 16.3 & 17.5 & \cellcolor{maincolor!12}1.2 & 30.53 & 31.93 & \cellcolor{maincolor!1}1.4 \\
\hline
GPT-3.5-Turbo & 29.45 & 43.12 & \cellcolor{maincolor!80}13.67 & 34.84 & 47.52 & \cellcolor{maincolor!82}12.68 & 25.92 & 30.81 & \cellcolor{maincolor!86}4.89 & 36.78 & 45.1 & \cellcolor{maincolor!80}8.32 \\
GPT-4o-mini & 34.11 & \uline{48.69} & \cellcolor{maincolor!87}14.58 & 41.77 & 54.47 & \cellcolor{maincolor!82}12.7 & 33.68 & \uline{39.23} & \cellcolor{maincolor!100}5.55 & 39.63 & \uline{47.55} & \cellcolor{maincolor!75}7.92 \\
Gemini-v1.5-Pro & \textbf{36.61} & 47.23 & \cellcolor{maincolor!59}10.62 & \textbf{45.46} & \uline{56.84} & \cellcolor{maincolor!72}11.38 & \uline{33.79} & 36.55 & \cellcolor{maincolor!44}2.76 & \textbf{42.73} & 47.35 & \cellcolor{maincolor!38}4.62 \\
Llama-3.1-70B & \uline{35.63} & \textbf{51.85} & \cellcolor{maincolor!99}16.22 & \uline{44.6} & \textbf{59.62} & \cellcolor{maincolor!100}15.02 & \textbf{37.13} & \textbf{42.6} & \cellcolor{maincolor!98}5.47 & \uline{40.93} & \textbf{50.96} & \cellcolor{maincolor!100}10.03 \\
\\[-2ex]
\hline
\textit{MvP-full} (SOTA) & 50.49 & 55.97 & \cellcolor{maincolor!22}5.48 & 59.31 & 63.98 & \cellcolor{maincolor!22}4.67 & 61.44 & 63.91 & \cellcolor{maincolor!38}2.47 & 57.05 & 62.63 & \cellcolor{maincolor!49}5.58 \\

\\[-2ex]
\hline
\rowcolor[HTML]{EFEFEF}
\multicolumn{13}{l}{\cellcolor[HTML]{EFEFEF}\textit{\textbf{Ensemble}}} \\

Paraphrase & 24.86 & 27.85 & \cellcolor{maincolor!4}2.99 & 34.75 & 36.46 & \cellcolor{maincolor!0}1.71 & 8.5 & 9.07 & \cellcolor{maincolor!0}0.57 & 24.6 & 25.91 & \cellcolor{maincolor!0}1.31 \\
DLO & 34.58 & 38.78 & \cellcolor{maincolor!13}4.2 & 41.31 & 46.22 & \cellcolor{maincolor!24}4.91 & 12.52 & 13.36 & \cellcolor{maincolor!5}0.84 & 37.6 & 38.99 & \cellcolor{maincolor!1}1.39 \\
MvP & 35.77 & 38.11 & \cellcolor{maincolor!0}2.34 & 44.85 & 47.25 & \cellcolor{maincolor!5}2.4 & 16.36 & 17.49 & \cellcolor{maincolor!11}1.13 & 34.27 & 35.55 & \cellcolor{maincolor!0}1.28 \\ 
\hline
GPT-3.5-Turbo & 35.64 & 48.9 & \cellcolor{maincolor!78}13.26 & 43.28 & 56.73 & \cellcolor{maincolor!88}13.45 & 30.88 & 34.39 & \cellcolor{maincolor!59}3.51 & 43.92 & 51.65 & \cellcolor{maincolor!73}7.73 \\
GPT-4o-mini & 40.36 & \uline{52.8} & \cellcolor{maincolor!72}12.44 & 46.59 & 59.59 & \cellcolor{maincolor!84}13.0 & 38.02 & \uline{42.03} & \cellcolor{maincolor!69}4.01 & \uline{47.55} & 52.96 & \cellcolor{maincolor!47}5.41 \\
Gemini-v1.5-Pro & \textbf{40.88} & 51.75 & \cellcolor{maincolor!60}10.87 & \textbf{49.44} & \uline{61.01} & \cellcolor{maincolor!74}11.57 & \uline{38.07} & 40.33 & \cellcolor{maincolor!34}2.26 & \textbf{49.61} & \uline{55.23} & \cellcolor{maincolor!49}5.62 \\
Llama-3.1-70B & \uline{40.41} & \textbf{56.74} & \cellcolor{maincolor!100}16.33 & \uline{48.99} & \textbf{63.97} & \cellcolor{maincolor!99}14.98 & \textbf{41.06} & \textbf{46.18} & \cellcolor{maincolor!91}5.12 & 46.0 & \textbf{55.8} & \cellcolor{maincolor!97}9.8 \\
\\[-2ex]
\hline
\textit{MvP-full} (SOTA) & 51.74 & 57.03 & \cellcolor{maincolor!21}5.29 & 59.75 & 64.15 & \cellcolor{maincolor!20}4.4 & 62.81 & 65.28 & \cellcolor{maincolor!38}2.47 & 59.02 & 64.67 & \cellcolor{maincolor!50}5.65 \\

\bottomrule
\end{tabular}%
}
\caption{
Comparison of average F1 scores between the original GTs (Orig.) and our expanded sets (Ours), highlighting the discrepancy between them ($\Delta$).
T5-based models, \ie Parapharse~\cite{zhang-2021-asqp-paraphrase}, DLO~\cite{hu-etal-2022-dlo-template}, MvP~\cite{gou-etal-2023-mvp}, are transfer learned from ASTE to our 20-shot examples. 
LLMs, specifically GPT-3.5-Turbo, GPT-4o-mini, Gemini-v1.5-Pro, Llama-3.1-70B-Instruct, generate predictions based on the same 20-shot in-context learning. 
The \textbf{best scores} are bolded, and the \uline{second-best scores} are underlined.
The state-of-the-art (SOTA) model, \ie \textit{full-finetuned MvP (MvP-full)}, is included for reference.
Note that the gap between the scores of the SOTA model and others decreases when evaluated on our GT set compared to the original set.
}
\label{tab:main-ensemble}
\end{table*}

\subsection{Dataset Validity}
\label{sec:dataset-validity}

To assess the validity of our expanded evaluation sets, we conduct human evaluations on our newly generated GTs.
Specifically, we randomly sample 80 examples from each dataset.
Table~\ref{tab:user-study-n-p} shows the number of quadruples for prediction and GT sets.
Three human evaluators are asked to determine whether our expanded GTs are valid considering the given sentence.
Before the evaluation, the evaluators were trained on the task and ensured to have a comprehensive understanding of the ABSA task.
The final human validity result was determined by a majority vote.
Details of the human evaluation, including the user interface sample, are provided in Appendix~\ref{appendix:details-human-agreement}.

As illustrated in Table~\ref{tab:validity-agreement}, our human evaluation reveals a high level of validity, with all percentages exceeding 90\%.
We also calculate the difference in the model F1 score between our test set and the human-filtered test set for the sampled examples.
Notably, there is an average 0.53\% F1 score difference between the two GT sets, implying the high quality of our new evaluation set. Detailed score is described in Table~\ref{tab:ours-oracle-diff}.

\subsection{Alignment with Human Evaluator}
\label{sec:human-agreement}

To demonstrate that evaluations with our expanded GTs align more closely with human judgments than those on the original GTs, we conduct human evaluations on model-predicted outputs and compare the alignment with our and the original GT sets. 
We randomly sample 80 test examples from each of the four datasets and extract predictions for two models: MvP and GPT-3.5-Turbo, which are consistent with the previous validity experiment. 
Three human annotators, provided with task descriptions and 20 examples identical to those given to the models, judge whether each prediction is appropriate for a given sentence and task. 
The final human decision is determined by a majority vote.

Inter-annotator agreement (IAA) is measured between human evaluators and the two GT sets (Origin vs. Humans and Ours vs. Humans) using Cohen's Kappa ($\kappa$) \citep{cohen1960coefficient} and Kendall Tau ($\tau$) \citep{kendall1938new}~\footnote{Pearson correlation is omitted due to its identical score to Kendall Tau, given the limited sample size.}.
Table~\ref{tab:human-agreement} shows that our new test set achieves higher IAA with human evaluators than the existing test set in most cases, demonstrating that our pipeline is more human-like than evaluations than the original set. 
MvP exhibits higher human agreement than other models, likely due to clearer distinctions between correct and incorrect predictions.
Also, the Fleiss' kappa\footnote{Fleiss' kappa was used to measure agreement among three annotators, as it is more appropriate than Cohen's kappa for more than two annotators.} among human annotators are 41.2, 44.3, 45.8, and 49.6 for ASQP-R15, R16, ACOS-Laptop, and Rest, respectively. 
This moderate agreement among annotators indicates a valid evaluation.
Based on these results, our expanded GT sets demonstrate improved alignment with human judgments, supporting the effectiveness of our approach in enhancing ABSA evaluation.

\subsection{Re-evaluate Model Performance}
\label{sec:model-performance}

We evaluate diverse ABSA approaches, comprising three T5-based models and four LLMs, on our expanded GT set using test examples from four datasets: ASQP-R15, R16, ACOS-Laptop, and Rest.
Data statistics are described in Table~\ref{tab:dataset-statistics}.
The scores are F1 using the exact match measurement.
Since the prediction order of the elements of each quadruple highly matters in accuracy~\cite{hu-etal-2022-dlo-template, gou-etal-2023-mvp}, all models predict 24 different orders. 
Naive scores are summarized in their average values.
We report the results using an ensemble alongside the naively predicted results.
The prediction in the ensemble is chosen if the prediction appears at least three times in top-5 order or seed.
The scores for all orders are in Appendix~\ref{sec:append-exp-results}.

As shown in Table~\ref{tab:main-ensemble}, our experimental results reveal significant performance improvements in ABSA tasks when assessing LLMs using our GT sets. 
LLM-based models demonstrate an average F1 score improvement of 9.8 percentage points, compared to 2.3 for T5-based models.
This substantial disparity suggests that conventional GT sets may have underestimated LLM performance in ABSA tasks due to their limited diversity of surface forms.
Notably, Llama-3.1 advances from second to first place in three out of four datasets under our new evaluation method, surpassing the initially top-performing Gemini.

The ensemble approach consistently outperforms naive methods, with Llama-3.1 achieving results within 1\%p of the SOTA model on the ASQP dataset. 
Interestingly, LLM performance varies between the ACOS and ASQP datasets, which differ in handling implicit opinion terms. 
The ASQP dataset, excluding ``null'' cases for implicit opinions, allows LLMs to demonstrate particular efficacy in extracting explicit opinion terms. 
This indicates that for tasks with explicit opinion expressions, LLMs can achieve performance levels comparable to specialized SOTA models. 
These findings collectively suggest that the potential of LLMs in ABSA tasks may have been previously underestimated, especially in scenarios with well-defined opinion expressions.

\begin{table}[t]
\small
\centering
\resizebox{\columnwidth}{!}{%
\begin{tabular}{c|c|c|ccc|c}
\toprule
\multicolumn{2}{c|}{} & \textbf{Orig.} & \textbf{\narrow} & \textbf{\widen} & \textbf{Filter\textsuperscript{*}} & \textbf{Ours} \\
\midrule
\multirow{2}{*}{ASQP-R15} & A & 795 & $+$212 & $+$679 & $-$313 & $+$578 \\
 & O & 1014 & $+$217 & $+$1480 & $-$704 & $+$993 \\
\hline
\multirow{2}{*}{ASQP-R16} & A & 799 & $+$203 & $+$732 & $-$346 & $+$589 \\
 & O & 799 & $+$208 & $+$1034 & $-$437 & $+$805 \\
\hline
\multirow{2}{*}{ACOS-Lap} & A & 1161 & $+$158 & $+$1117 & $-$477 & $+$798 \\
 & O & 1161 & $+$163 & $+$1415 & $-$402 & $+$1176 \\
\hline
\multirow{2}{*}{ACOS-Rest} & A & 916 & $+$97 & $+$791 & $-$300 & $+$588 \\
 & O & 916 & $+$115 & $+$1393 & $-$614 & $+$894 \\
\bottomrule
\end{tabular}%
}
\caption{
Number of terms added or removed at each step of \method for aspect (A) and opinion (O).
Our pipeline begins with the original terms (Orig.), generates new terms in \narrow and \widen, and eliminates terms through \judge and rule-based filtering (Filter*), resulting in a new GT set (Ours).
}
\label{tab:data-gen-statistics}
\end{table}

\begin{figure}[t]
\begin{center}
\includegraphics[width=\columnwidth]{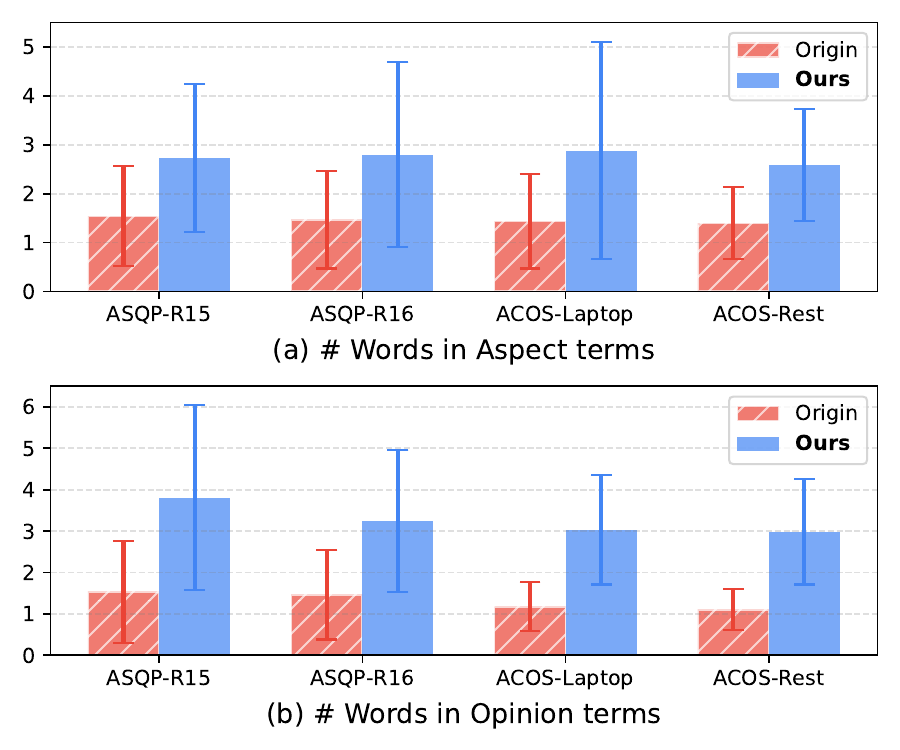}
\end{center}
\vspace{-3mm}
\caption{
Average and standard deviation of word counts in explicit aspect and opinion terms across the four datasets.
}
\label{fig:word-count}
\end{figure}

\section{Analysis}

\subsection{Statistics of Our GT set}
\label{sec:analysis-dataset-statistics}

\paragraph{Changes in each step}
Table~\ref{tab:data-gen-statistics} shows the number of terms added or removed at each pipeline step for aspect (A) and opinion (O) terms. 
Our final datasets contain, on average, 638 more aspect and 967 more opinion terms than the original GT set.
\widen generates more terms in both aspect and opinion than \narrow step.
This is natural because \widen extracts terms from anywhere in the sentence, while \narrow is limited to the original GT term.

\paragraph{Number of words in a term}
We compute the number of words in aspect and opinion terms using the NLTK word tokenizer \citep{bird-loper-2004-nltk}, excluding implicit cases where the term is ``null.''
Figure~\ref{fig:word-count} illustrates the mean and standard deviation of the word counts for both the original and our test sets. 
Our terms exhibit higher word counts in all cases, suggesting a broader range of surface forms than the original dataset.
The observed increases in standard deviation further suggest that our dataset selectively adds words where necessary, rather than indiscriminately increasing word count across all terms.

\subsection{Ablation Study}
\label{sec:analysis-ablation}

We validate the role of \narrow, \widen, and \judge steps in expanding the existing GT set.
Following the experimental setup in Section~\ref{sec:human-agreement}, we analyze the agreement with human evaluation for each step.

Table~\ref{tab:ablation-study-IAA-gpt} shows the number of expanded quadruples for the sampled 80 examples with human ratings, along with the changes in IAA for each step.
The \narrow and \widen steps generally improve IAA, except for the \widen step in ACOS-Laptop.
This result implies that each of these processes helps align existing GT with human evaluation criteria.
On the other hand, the IAA drop in the \widen process of ACOS-Laptop is attributed to an issue with categorical values, which our extension does not handle.
This can be understood as a limitation of our work since our model does not consider the variability of category and sentiment.
In the \judge process, while the number of quadruples decreases by approximately 20\% on average, IAA improves or remains the same, indicating the judging process removes irrelevant quadruples that do not affect human alignment and fairly improves the dataset quality.

\subsection{Comparison with SOTA Model}
In Table~\ref{tab:main-ensemble}, Llama-3.1-70B shows high accuracy, approaching the SOTA performance of MvP in our evaluation set.
Especially in the ensemble setting, the difference is merely 0.2\%p.
To further analyze the capability of Llama, we increase the number of demonstration examples while maintaining the same experimental setup as in Table~\ref{tab:main-ensemble}.

As illustrated in Table~\ref{tab:llama-few-shot}, Llama outperforms MvP in the 40-shot ensemble setting when evaluated on our GT set.
This finding implies that Llama has been apparently underestimated in the conventional single-answer exact match evaluation approach.
Our observation aligns with previous research in RE tasks where human evaluation demonstrates the near-SOTA performance of LLMs~\cite{wadhwa-etal-2023-revisiting}.
Notably, these prediction results derive from a simple few-shot ICL setting. 
We expect that applying advanced techniques, such as example selection and chain-of-thoughts prompting, would yield results comparable to or exceeding current SOTA performance.
We leave this exploration as our future work.


\begin{table}[t]
\resizebox{\columnwidth}{!}{%
\begin{tabular}{l||cc|cc|cc|cc}
\toprule
\multicolumn{1}{c||}{\multirow{2}{*}{\textbf{Models}}} & \multicolumn{2}{c|}{\textbf{ASQP-R15}} & \multicolumn{2}{c|}{\textbf{ASQP-R16}} & \multicolumn{2}{c|}{\textbf{ACOS-Lap}} & \multicolumn{2}{c}{\textbf{ACOS-Rest}} \\
\multicolumn{1}{c||}{} & \textbf{\#Q} & \textbf{IAA} & \textbf{\#Q} & \textbf{IAA} & \textbf{\#Q} & \textbf{IAA} & \textbf{\#Q} & \textbf{IAA} \\
\midrule
\textbf{Orig.} & 128 & 8.3 & 110 & 26.3 & 114 & 12.8 & 132 & 23.5 \\
\textbf{+\narrow} & 215 & \textbf{11.9} & 185 & \textbf{27.1} & 145 & \textbf{16.1} & 166 & \textbf{25} \\
\textbf{+\widen} & 659 & \textbf{18} & 476 & \textbf{37} & 503 & 15 & 609 & \textbf{31.1} \\
\midrule
\textbf{Ours} & 546 & \textbf{18} & 402 & \textbf{48.9} & 369 & \textbf{15} & 487 & \textbf{31.1}
\\ \bottomrule
\end{tabular}%
}
\caption{
Inter-annotator agreement (IAA) measured by Cohen's kappa ($\kappa$) between human annotators and ground truth (GT) sets at each step of processing GPT-3.5-Turbo's predictions.
The number of GT quadruples (\#Q) increases during generation steps and decreases after filtering. 
Scores are \textbf{bolded} if they equal or exceed those from the previous step.
}
\label{tab:ablation-study-IAA-gpt}
\end{table}

\begin{table}[t]
\centering
\resizebox{.9\columnwidth}{!}{%
\begin{tabular}{l|c|ccc}
\toprule
\textbf{Models} & \textbf{\#Train} & \textbf{Orig.} & \textbf{Ours} & \textbf{$\Delta$} \\ \hline
\multicolumn{5}{l}{\cellcolor[HTML]{EFEFEF}\textit{\textbf{Naive}}} \\
\textit{MvP-full (SOTA)} & 1,264 & \textbf{59.31} & \textbf{63.98} & 4.67 \\ \hline
\multirow{2}{*}{Llama-3.1-70B} & 20 & 44.6 & 59.62 & 15.02 \\
 & 40 & \uline{46.33} & \uline{62.14} & 15.81 \\ 
\\[-2.5ex]
\hline
\multicolumn{5}{l}{\cellcolor[HTML]{EFEFEF}\textit{\textbf{Ensemble}}} \\
\textit{MvP-full (SOTA)} & 1,264 & \textbf{59.75} & \uline{64.15} & 4.4 \\ \hline
\multirow{2}{*}{Llama-3.1-70B} & 20 & 48.99 & 63.97 & 14.98 \\
 & 40 & \uline{50.09} & \textbf{66.42} & 16.33 \\
\bottomrule
\end{tabular}%
}
\caption{Comparison of full-finetuned MvP, \ie the SOTA model, and Llama-3.1-70B-Instruct in few-shot ICL setting on ASQP-Rest16 dataset. The scores are averaged F1 exact match scores in 24 orders.
}
\label{tab:llama-few-shot}
\end{table}

\section{Conclusion}
We present \method, a novel approach for the ABSA task that addresses the limitations of traditional evaluation methods by accounting for the diversity of surface forms in aspect and opinion terms.
By expanding the ground truth sets to include valid variations that preserve original meanings, our method aligns more closely with human judgments. 
Experimental results show that LLMs significantly outperform T5 models under the proposed evaluation scheme, suggesting that LLMs' capabilities in span extraction tasks have been underestimated due to restrictive evaluation practices. 
This underscores the importance of considering the diversity of expression in the LLM era for fair and accurate assessments. 
While our study focuses on ABSA, the principles underlying \method are applicable to other NLP tasks involving span extraction, such as named entity recognition, relation extraction, and event extraction.
Owing to the flexibility of our framework, we anticipate researchers in NLP tasks, especially for information extraction tasks such as named entity recognition or relation extraction, which include span extraction tasks.

\section*{Limitation}
Our study performed the quadruple prediction task between the various tasks of ABSA, which included tuple or triplet predictions.
However, such tasks exist as part of quadruples, so there is plenty of room for our approach to be utilized.
Also, only explicit mention in the sentence is covered in our experiments.
This is because the implicit cases tagged in ``null'' make it difficult to extract diverse surface forms where the meaning of ``null'' is hidden in the sentence.

\section*{Ethics Statement}
All human evaluators voluntarily agreed to participate in this study.
We informed them that their responses would be anonymized and securely protected.
Evaluators were free to stop the survey at any time.
Compensation was more than adequate.
As our survey and dataset do not contain any sensitive or harmful content, we do not anticipate any negative ethical impacts from our research.
Also, we utilize ChatGPT and Claude as a writing assistant and Copilot as a coding assistant.

\section*{Acknowledgements}
This work was supported by Institute for Information \& communications Technology Promotion(IITP) grant funded by the Korea government(MSIT) (No.RS-2019-II190075 Artificial Intelligence Graduate School Program(KAIST)), Artificial intelligence industrial convergence cluster development project funded by the Ministry of Science and ICT(MSIT, Korea) \& Gwangju Metropolitan City, and the National Research Foundation of Korea (NRF) grant funded by the Korea government (MSIT) (No. NRF-2022R1A2B5B02001913).
Lastly, we sincerely appreciate to our DAVIAN Lab people, sharing insightful feedback and discussions.


\bibliography{custom}

\newpage
\appendix
\part*{Appendix}

\section{Experimental Details}
\label{sec:append-implementation-detail}

\subsection{Dataset Statistics}

Table~\ref{tab:dataset-statistics} shows statistics of dataset utilized in our experiments, \ie ASQP~\cite{zhang-2021-asqp-paraphrase} and ACOS~\cite{cai-2021-acos}.

\begin{table}[h]
\centering
\resizebox{.8\columnwidth}{!}{%
\begin{tabular}{l|rrr}
\toprule
\textbf{Datasets} & \multicolumn{1}{l}{\textbf{\# Train}} & \multicolumn{1}{l}{\textbf{\# Dev.}} & \multicolumn{1}{l}{\textbf{\# Test}} \\ \midrule
ASQP-R15 & 834 & 209 & 537 \\
ASQP-R16 & 1,264 & 316 & 544 \\
ACOS-Laptop & 2,934 & 326 & 816 \\
ACOS-Rest & 1,530 & 171 & 583 \\
\bottomrule
\end{tabular}%
}
\caption{Statistics for each dataset. The number of examples for train, dev, and test sets.}
\label{tab:dataset-statistics}
\end{table}

\subsection{Implementation Details}
For Llama models, we utilize NVIDIA H100 80GB HBM3.
To fine-tune T5 models, we used NVIDIA GeForce RTX 3090 and NVIDIA A100-PCIE-40GB.
Table~\ref{tab:construct-cost} shows the total cost to construct our GT set.
The dataset was constructed in May 2024.

\begin{table}[h]
\centering
\resizebox{\columnwidth}{!}{%
\begin{tabular}{c|cc|cc|cc|cc}
\toprule
\multirow{2}{*}{\textbf{Dataset}} & \multicolumn{2}{|c}{\textbf{ASQP-R15}} & \multicolumn{2}{|c}{\textbf{ASQP-R16}} & \multicolumn{2}{|c}{\textbf{ACOS-L}} & \multicolumn{2}{|c}{\textbf{ACOS-R}} \\
 & \textbf{A} & \textbf{O} & \textbf{A} & \textbf{O} & \textbf{A} & \textbf{O} & \textbf{A} & \textbf{O} \\
\midrule
\narrow & 5.6 & 7.1 & 5.9 & 6.9 & 8.6 & 7.9 & 6.6 & 6.9 \\
\widen & 7.4 & 9.7 & 7.6 & 9.6 & 10.3 & 8.8 & 7.7 & 7.6 \\
\judge & 16.3 & 30.4 & 16.3 & 21.5 & 21.7 & 27.2 & 16.1 & 26.4 \\
\hline
Sum. element & 29.3 & 47.2 & 29.8 & 38 & 40.6 & 43.9 & 30.4 & 40.9 \\
\textbf{Sum. dataset} & \multicolumn{2}{|c}{76.5} & \multicolumn{2}{|c}{67.8} & \multicolumn{2}{|c}{84.5} & \multicolumn{2}{|c}{71.3} \\
\bottomrule
\end{tabular}%
}
\caption{Total cost (\$) to construct expanded GT sets for each step, element (aspect (A) and opinion (O)), and dataset.}
\label{tab:construct-cost}
\end{table}

\section{Details of Human Evaluation}
\label{appendix:details-human-agreement}

In this section, we describe the full results and the experimental details of Section~\ref{sec:dataset-validity} and Section~\ref{sec:human-agreement}, which include human annotations.
Three authors volunteered to participate in this study.
We used the four ASQP datasets and randomly selected 80 samples in test sets where the seed was 42.

\begin{table*}[ht]
\resizebox{\textwidth}{!}{%
\begin{tabular}{l||ccc|ccc|ccc|ccc}
\toprule
\textbf{Datasets} & \multicolumn{3}{c|}{\textbf{ASQP-R15}} & \multicolumn{3}{c|}{\textbf{ASQP-R16}} & \multicolumn{3}{c|}{\textbf{ACOS-Lap}} & \multicolumn{3}{c}{\textbf{ACOS-Rest}} \\ 
\textbf{Models} & \multicolumn{1}{l}{\textbf{Ours}} & \multicolumn{1}{l}{\textbf{Oracle}} & \multicolumn{1}{l|}{\textbf{Diff}} & \multicolumn{1}{l}{\textbf{Ours}} & \multicolumn{1}{l}{\textbf{Oracle}} & \multicolumn{1}{l|}{\textbf{Diff}} & \multicolumn{1}{l}{\textbf{Ours}} & \multicolumn{1}{l}{\textbf{Oracle}} & \multicolumn{1}{l|}{\textbf{Diff}} & \multicolumn{1}{l}{\textbf{Ours}} & \multicolumn{1}{l}{\textbf{Oracle}} & \multicolumn{1}{l}{\textbf{Diff}} \\ \midrule
\textit{MvP-full} & 39.25 & 38.49 & 0.76 & 46.96 & 46.96 & 0 & 19.82 & 19.82 & 0 & 25.28 & 25.28 & 0 \\
GPT-3.5-Turbo & 39.27 & 37.82 & 1.45 & 41.43 & 41.43 & 0 & 31.02 & 30.2 & 0.82 & 39.02 & 38.33 & 0.69 \\
GPT-4o-mini & 46.32 & 44.85 & 1.47 & 59.41 & 59.41 & 0 & 39.15 & 38.3 & 0.85 & 44.12 & 42.65 & 1.47 \\
Gemini-pro & 49.06 & 49.06 & 0 & 62.07 & 62.07 & 0 & 45.81 & 45.81 & 0 & 43.08 & 42.31 & 0.77 \\
Llama-3.1-70B-Instruct & 49.25 & 47.76 & 1.49 & 57.63 & 57.63 & 0 & 39.82 & 39.82 & 0 & 46.87 & 46.09 & 0.78 \\ \midrule
Average & 44.63 & 43.60 & 1.03 & 53.50 & 53.50 & 0.00 & 35.12 & 34.79 & 0.33 & 39.67 & 38.93 & 0.74 \\ \bottomrule
\end{tabular}%
}
\caption{Difference in the model F1 score (\%) between our test set (\textbf{Ours}) and the human-filtered test set (\textbf{Oracle}) for the sampled examples.}
\label{tab:ours-oracle-diff}
\end{table*}

\begin{figure}[ht]
\begin{center}
\fbox{\includegraphics[width=0.95\columnwidth]{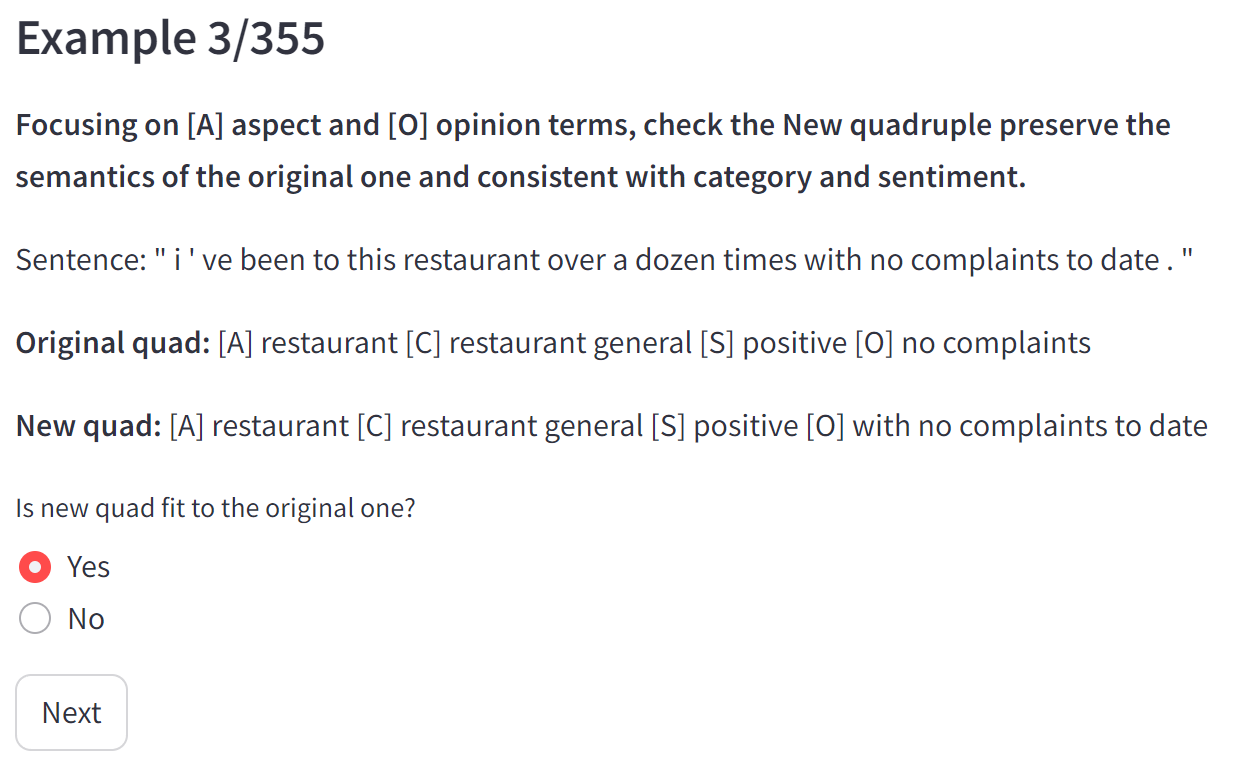}}
\end{center}
\caption{Dataset validity study UI example on ACOS-Rest dataset in Section~\ref{sec:dataset-validity}.}
\label{fig:appx-ui-validity-example}
\end{figure}

\begin{figure}[ht]
\begin{center}
\fbox{\includegraphics[width=0.95\columnwidth]{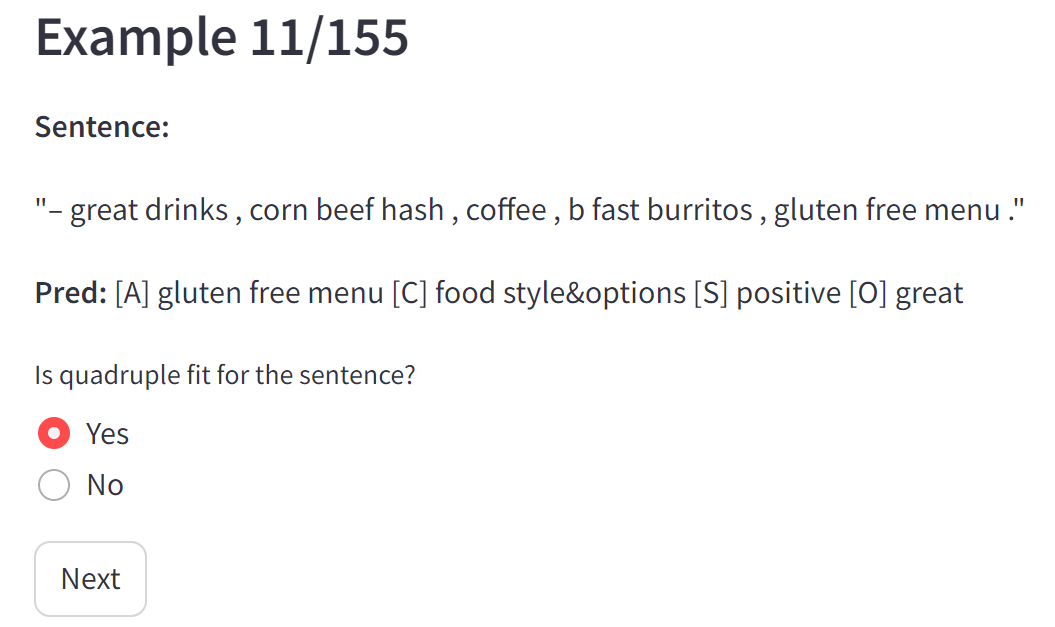}}
\end{center}
\caption{Human evaluation study UI example on ACOS-Rest dataset.}
\label{fig:appx-ui-example}
\end{figure}

\subsection{Dataset Validity}
In this experiment, we asked human annotators to validate whether the new GT quadruple is correct with the given sentence.
The user interface is shown in Figure~\ref{fig:appx-ui-validity-example}.
Also, Table~\ref{tab:ours-oracle-diff} describes the model performance difference between our GT set and the human annotator's oracle GT.

\subsection{Human evaluation}
Human annotators evaluated the predicted quadruples from two models: MvP and GPT-3.5-Turbo. 
The annotators were provided with a sentence and a predicted quadruple and asked to tag whether the predicted quadruple was correct or not. The user interface is shown in Figure~\ref{fig:appx-ui-example}.

\section{Further Analysis}

\subsection{Examples of New GT set}

Table~\ref{tab:examples} illustrates the aspect and opinion terms deemed acceptable in true positive cases within our evaluation set. 
In the first example, the prediction that ``sake'' is ``successfully easing'' is assessed as correct in our revised ground truth (GT) set, whereas the original GT would have classified it as incorrect. 
This example demonstrates that the newly accepted term aligns semantically with the original annotation, justifying its inclusion as a valid prediction.

\subsection{Ablation study}
As described in Section~\ref{sec:analysis-ablation}, we measure IAA between humans and variants of our GT sets for each step. Table~\ref{tab:appendix-ablation-full} shows the full result.

\section{Prompts}
\label{sec:Prompts}

\subsection{System Prompt}
\label{sec:append-system-prompt}

Here, we show the full system prompts for our dataset expansion pipeline and model prediction.
For dataset construction, three steps are done independently for each aspect and opinion term:
Table~\ref{tab:narrow_prompt} for \narrow step,
Table~\ref{tab:widen_prompt} for \widen step, Table~\ref{tab:judge_prompt_aspect} for \judge step in aspect term, 
Table~\ref{tab:judge_prompt_opinion} for \judge step in opinion term.
On the other hand, to get the prediction of diverse LLMs, we input the same 20-shot, as shown in Table~\ref{tab:model_pred_prompt}.

\subsection{Examples in \method}
In \narrow and \widen, 5-shot ICL is used. The example is described in Table~\ref{tab:icl_prompt}.
In \judge step, to exercise caution in judgment, we utilize 5-shot CoT prompting. Table~\ref{tab:llm_demo_example} shows the demonstration example, and Table~\ref{tab:judge_ouptut} is the real decision of LLM classifying the validness of the new term.

\subsection{Demonstration Examples}
\label{sec:append-demo}
20 examples are selected by heuristics considering the balance of implicit cases, category, and sentiment. Table~\ref{tab:demonstration_examples} shows the selected examples for the ACOS-Rest dataset.




\begin{table*}[t]
\centering
{\normalsize
\resizebox{\textwidth}{!}{%
\begin{tabular}{lcll}
\toprule
\multicolumn{1}{c}{\textbf{Original}} &     \contour{black}{$\rightarrow$}
                & \multicolumn{1}{c}{\textbf{Ours}}       & \multicolumn{1}{c}{\textbf{Generated Step}}        \\ \midrule
\multicolumn{4}{l}{\begin{tabular}[c]{@{}l@{}}\textbf{Sentence}: the sake ’ s complimented the courses very well and is successfully easing me into the sake world .\end{tabular}} \\
\\[-2ex]
 \quad \textbf{{[}A{]}} sake ’ s \textbf{{[}O{]}} successfully                   & \contour{black}{$\rightarrow$} & \textbf{{[}A{]}} \inbox{sake} \textbf{{[}O{]}} \outbox{successfully easing} & \textbf{{[}A{]}} \inbox{\narrow} \textbf{{[}O{]}} \outbox{{\widen}} \\ \midrule
\multicolumn{4}{l}{\begin{tabular}[c]{@{}l@{}}\textbf{Sentence}: the lemon chicken tasted like sticky sweet donuts and the honey walnut prawns , the few they actually give you ... were not good .\end{tabular}}            \\
\\[-2ex]
\quad \textbf{{[}A{]}} lemon chicken \textbf{{[}O{]}} sticky sweet                  & \contour{black}{$\rightarrow$} & \textbf{{[}A{]}} \originbox{lemon chicken} \textbf{{[}O{]}} \outbox{tasted like sticky sweet donuts} & \textbf{{[}A{]}} \originbox{\textsc{Original}} \textbf{{[}O{]}} \outbox{{\widen}} \\\midrule
\multicolumn{4}{l}{\begin{tabular}[c]{@{}l@{}}\textbf{Sentence}: thus far , i ' ve loaded a number of games , my password vault , several productivity apps , skype , spotify and some network \\ scanning android apps and they ' ve all worked very well .\end{tabular}}            \\
\\[-2ex]
\quad \textbf{{[}A{]}} android apps \textbf{{[}O{]}} well                   & \contour{black}{$\rightarrow$} & \textbf{{[}A{]}} \outbox{network scanning android apps} \textbf{{[}O{]}} \outbox{very well} & \textbf{{[}A{]}} \outbox{\widen} \textbf{{[}O{]}}\outbox{\widen} \\ \midrule
\multicolumn{4}{l}{\begin{tabular}[c]{@{}l@{}}\textbf{Sentence}:  the side of potatoes is to die for , as is the labne ( yogurt dip ) .\end{tabular}}            \\
\\[-2ex]
\quad \textbf{{[}A{]}} labne ( yogurt dip ) \textbf{{[}O{]}} die for                   & \contour{black}{$\rightarrow$} & \textbf{{[}A{]}} \inbox{labne} \textbf{{[}O{]}} \originbox{die for} & \textbf{{[}A{]}}\inbox{\narrow} \textbf{{[}O{]}}\originbox{\textsc{Original}} \\ 
\bottomrule
\end{tabular}%
}
}
\caption{
Four examples have been included as true positives in our GT set.
Each row contains examples sourced from the ASQP-R15, ASQP-R16, ACOS-Laptop, and ACOS-Rest datasets.
The generated step for each aspect and opinion term is explained separately.
}
\label{tab:examples}
\end{table*}

\section{LLMs Performance}
\label{sec:append-exp-results}

Full experimental results in Table~\ref{tab:main-ensemble} are shown in three tables for three different LLMs as follows: 
Table~\ref{tab:acos-asqp-gpt} for GPT-3.5-Turbo, Table~\ref{tab:acos-asqp-gpt-4o-mini} for GPT-4o-mini, Table~\ref{tab:acos-asqp-gemini} for Gemini-v1.5-Pro, and Table~\ref{tab:acos-asqp-llama3.1} for Llama-3.1-70B-Instruct.

\begin{table*}[t]
\resizebox{\textwidth}{!}{%
\begin{tabular}{l||ccc|ccc|ccc|ccc}
\toprule
 & \multicolumn{3}{c|}{\textbf{ASQP-R15}} & \multicolumn{3}{c|}{\textbf{ASQP-R16}} & \multicolumn{3}{c|}{\textbf{ACOS-Lap}} & \multicolumn{3}{c}{\textbf{ACOS-Rest}} \\
\multirow{-2}{*}{} & \textbf{\#GT} & \textbf{MvP} & \textbf{GPT} & \textbf{\#GT} & \textbf{MvP} & \textbf{GPT} & \textbf{\#GT} & \textbf{MVP} & \textbf{GPT} & \textbf{\#GT} & \textbf{MvP} & \textbf{GPT} \\
\midrule
\textbf{Orig.} & 128 & 33.4/42.3 & 8.3/18.8 & 110 & 56.6/61.8 & 26.3/38.9 & 114 & 50.5/58.1 & 12.8/22.5 & 132 & 47.3/54.5  & 23.5/33.9 \\
\textbf{+\narrow} & 215 & \textbf{39.5/47.4} & \textbf{11.9/23.4} & 185 & \textbf{60.9/65.3} & \textbf{27.1/39.6} & 145 & 50.5/58.1 & \textbf{16.1/26.2} & 166 & 47.3/54.5  & \textbf{25/35.3} \\
\textbf{+\widen} & 659 & \textbf{40.6/48.2} & \textbf{18/29.8} & 476 & 60.9/65.3 & \textbf{37/46.5} & 503 & \textbf{52.7/59.8} & 15/23.7 & 609 & 47.3/54.5 & \textbf{31.1/40.6} \\
\midrule
\textbf{Ours} & 546 & {39.5/47.4} & 18/29.8 & 402 & {59.5/64.1} & \textbf{38.7/48.9} & 369 & 52.7/59.8 & 15/23.7 & 487 & 47.3/54.5 & 31.1/40.6
\\ \bottomrule
\end{tabular}%
}
\caption{Agreement scores of Cohen's kappa and Kendal tau ($\kappa$ / $\tau$). Each row shows the accumulation of each module.
The models are MvP and GPT-3.5-Turbo, respectively. Scores are \textbf{bolded} when the score increases to the previous step.}
\label{tab:appendix-ablation-full}
\end{table*}

\begin{table*}[ht]
    \begin{subtable}{\linewidth}
        \begin{tcolorbox}[
            colback=orangebg,
            colframe=black,
            fontupper=\footnotesize
        ]
        Given an input sentence, Sentiment, Opinion, Categroy, and \textbf{target Aspect terms}, extract expressions that narrow the span of the aspect term. The new expressions must be confined within the original aspect term and adhere to the following criteria:

        \begin{enumerate}[topsep=0pt,itemsep=-1ex,partopsep=1ex,parsep=1ex]
            \item Remain relevant to the given aspect term without altering its original meaning.
        
        \item Exclude any unnecessary words or spaces.
        
        \item Correct any typos if present and resolve contraction if present.
        
        \item Revert to the original expression if narrowing proves challenging.
        
        \item Ensure the expression exists exactly as it appears in the given sentence.
        
        \item Keep the aspect term and opinion term distinct and independent.
        \end{enumerate}
        \end{tcolorbox}
    \end{subtable}

\begin{subtable}{\linewidth}
\begin{tcolorbox}[
    colback=bluebg,
    colframe=black,
    fontupper=\footnotesize
]
Given an input sentence, Category, Aspect, Sentiment, and \textbf{target Opinion terms}, extract expressions that narrow the span of the opinion term. The new expressions must be confined within the original opinion term and adhere to the following criteria:

\begin{enumerate}[topsep=0pt,itemsep=-1ex,partopsep=1ex,parsep=1ex]
    \item Be related to both the aspect and opinion terms. Maintain the original sentiment polarity; changes in sentiment by narrowing the span are not allowed.

    \item Correct any typos and resolve contractions if present.

\item Exclude any unnecessary words or spaces.

\item Return the original expression if reducing it proves difficult.

\item Ensure the expression exists verbatim in the given sentence.

\item Keep the aspect and opinion terms distinct and independent.
\end{enumerate}
\end{tcolorbox}

\end{subtable}
\caption{\narrow step system prompts for \aspect{Aspect} and \opinion{Opinion}, respectively.}
\label{tab:narrow_prompt}
\end{table*}
\begin{table*}[ht]
    \begin{subtable}{\linewidth}
        \begin{tcolorbox}[
            colback=orangebg,
            colframe=black,
            fontupper=\footnotesize
        ]
        Given an input sentence, Sentiment, Opinion, Categroy, and \textbf{target Aspect terms}, extract various expressions that extend the span of the aspect term. The new expressions should be formed by adding surrounding words and must meet the following criteria:

\begin{enumerate}[topsep=0pt,itemsep=-1ex,partopsep=1ex,parsep=1ex]
    \item Be related to the given aspect term.
    \item Include neighboring words around the aspect term to form a new expression.
    \item Should not overlap with the original opinion term [O].
\item While expanding the aspect term, avoid incorporating the entire sentence.
\item If it is not feasible to expand the expression, return only the original aspect term.
\item Ensure the new expression matches exactly as it appears in the input sentence.
\end{enumerate}
        \end{tcolorbox}
    \end{subtable}

\begin{subtable}{\linewidth}
\begin{tcolorbox}[
    colback=bluebg,
    colframe=black,
    fontupper=\footnotesize
]
Given an input sentence, Category, Aspect, Sentiment, and \textbf{target Opinion terms}, extract various expressions that extend the span of the opinion term. The new expressions should adhere to the following criteria:
\begin{enumerate}[topsep=0pt,itemsep=-1ex,partopsep=1ex,parsep=1ex]
    \item Be related to the given opinion term.
\item Should not overlap with the original aspect term [A].
\item Include neighboring words around the opinion term to form a new expression, while ensuring the new expression does not encompass the entire sentence.
\item If it is challenging to expand the expression, return only the original opinion term.
\item Ensure the sentiment polarity remains consistent; expanding the expression should not alter the given sentiment polarity.
\end{enumerate}
\end{tcolorbox}

\end{subtable}
\caption{\widen step system prompts for \aspect{Aspect} and \opinion{Opinion}, respectively.}
\label{tab:widen_prompt}
\end{table*}
\begin{table*}[ht]
        \begin{tcolorbox}[
            colback=orangebg,
            colframe=black,
            fontupper=\footnotesize
        ]
        You are tasked with assessing whether a newly created aspect term aligns with a given Ground Truth (GT) quadruple in aspect-based sentiment analysis (ABSA). Here's how to do it:

\begin{enumerate}
    \item Review the provided sentence and the GT quadruple, which includes:

    \begin{itemize}
        \item[-] Aspect Term (A): The specific word or phrase referring to an aspect in the sentence.
   
        \item[-] Opinion Term (O): The word or phrase expressing an opinion about the aspect.
   
        \item[-] Aspect Category (C): The category to which the aspect term belongs. Categories include:

        \begin{itemize}
        \setlength{\parskip}{0pt}
        \setlength{\itemsep}{0pt plus 1pt}
            \item Location General
            \item Food Prices
            \item Food Quality
            \item Food General
            \item Food Style\&Options
            \item Ambience General
            \item Service General
            \item Restaurant General
            \item Restaurant Prices
            \item Restaurant Miscellaneous
            \item Drinks Prices
            \item Drinks Quality
            \item Drinks Style\&Options
        \end{itemize}
        \item[-] Sentiment Polarity (S): The sentiment associated with the opinion, chosen from:
        \begin{itemize}
            \item Positive
            \item Neutral
            \item Negative
     \end{itemize}
    \end{itemize}

\item Determine the alignment based on the following criteria:

    \begin{enumerate}
        \item[1.] Aspect and Category Consistency:
        
      - The new aspect term must maintain the target object of the [A] aspect and the [C] category in the GT.
      
        \item[2.] Sentiment and Opinion Relevance:
        
      - The new aspect term must directly relate to the [S] sentiment and [O] opinion as the GT.
      
        \item[3.] Extractability:
        
      - The new aspect term must be directly taken from the sentence without adding new words or significantly rearranging existing ones.
      - Minor adjustments like unwinding contractions or fixing typos are allowed.

        \item[4.] Independency:
        
      - Each aspect and opinion term must be independent and not overlap.
      
      - The new aspect term must not contain the GT [O] opinion term.
    \end{enumerate}

\item Determining Validity:
   \begin{itemize}
       \item[-] If all criteria are met, the new term is "valid."
        \item[-] If any criterion is not met, the new term is "invalid."

   \end{itemize}

\item Providing Feedback:
    \begin{itemize}
        \item[-] Explain why a term was deemed valid or invalid based on the above criteria.
        \item[-] Specific feedback helps in understanding the decision.
   \end{itemize}
   \end{enumerate}
        \end{tcolorbox}
    \caption{System prompt of Judge step for \aspect{Aspect} terms, especially on ACOS-Rest dataset.}
\label{tab:judge_prompt_aspect}
\end{table*}

\begin{table*}
\begin{tcolorbox}[
    colback=bluebg,
    colframe=black,
    fontupper=\footnotesize
]
You are tasked with assessing whether a newly created opinion term aligns with a given Ground Truth (GT) quadruple in aspect-based sentiment analysis (ABSA). Here's how to do it:

\begin{enumerate}
    \item Review the provided sentence and the GT quadruple, which includes:
    \begin{itemize}
        \item[-] Aspect Term (A): The specific word or phrase referring to an aspect in the sentence.
        \item[-] Opinion Term (O): The word or phrase expressing an opinion about the aspect.
        \item[-] Aspect Category (C): The category to which the aspect term belongs. Categories include:
        \begin{itemize}
            \item Location General
            \item Food Prices
            \item Food Quality
            \item Food General
            \item Food Style\&Options
            \item Ambience General
            \item Service General
            \item Restaurant General
            \item Restaurant Prices
            \item Restaurant Miscellaneous
            \item Drinks Prices
            \item Drinks Quality
            \item Drinks Style\&Options
        \end{itemize}
        \item[-] Sentiment Polarity (S): The sentiment associated with the opinion, chosen from:
        \begin{itemize}
            \item Positive
            \item Neutral
            \item Negative
     \end{itemize}
    \end{itemize}
\item Determine the alignment based on the following criteria:
    \begin{enumerate}
        \item[1.] Aspect and Category Relevance:
        \begin{itemize}
            \item[-] The new opinion term must directly relate to the [A] aspect and the [C] category in the GT.
        \end{itemize}
        
        \item[2.] Sentiment and Opinion Consistency:
        \begin{itemize}
            \item[-] The new opinion term should maintain the same [S] sentiment polarity and [O] opinion as the GT.
        \end{itemize}
        
        \item[3.] Extractability:
        \begin{itemize}
            \item[-] The new opinion term must be directly taken from the sentence without adding new words or significantly rearranging existing ones.
            \item[-] Minor adjustments like unwinding contractions or fixing typos are allowed.
        \end{itemize}
        \item[4.] Independency:
        \begin{itemize}
            \item[-] Each aspect and opinion term must be independent and not overlap.
            \item[-] The new opinion term must not contain the GT [A] aspect term.
      \end{itemize}
    \end{enumerate}
\item Determining Validity:
    \begin{itemize}
        \item[-] If all criteria are met, the new term is "valid."
        \item[-] If any criterion is not met, the new term is "invalid."
    \end{itemize}
    
\item Providing Feedback:
\begin{itemize}
    \item[-] Explain why a term was deemed valid or invalid based on the above criteria.
   \item[-] Specific feedback helps in understanding the decision.
\end{itemize}
\end{enumerate}
\end{tcolorbox}
\caption{System prompt of Judge step for \opinion{Opinion} terms, especially on ACOS-Rest dataset.}
\label{tab:judge_prompt_opinion}
\end{table*}

\begin{table*}[ht]
    \begin{subtable}{\linewidth}
        \begin{tcolorbox}[
            colback=bluebg,
            colframe=black,
            fontupper=\footnotesize
        ]

        \textit{\#\# System Prompt}
        
        Given an input sentence, Category, Aspect, Sentiment, and \textbf{target Opinion terms}, extract expressions that narrow the span of the opinion term. The new expressions must be confined within the original opinion term and adhere to the following criteria:

\begin{enumerate}[topsep=0pt,itemsep=-1ex,partopsep=1ex,parsep=1ex]
    \item Be related to both the aspect and opinion terms. Maintain the original sentiment polarity; changes in sentiment by narrowing the span are not allowed.

    \item Correct any typos and resolve contractions if present.

\item Exclude any unnecessary words or spaces.

\item Return the original expression if reducing it proves difficult.

\item Ensure the expression exists verbatim in the given sentence.

\item Keep the aspect and opinion terms distinct and independent.
\end{enumerate}

\textit{\#\# Demonstration}

Input sentence: "the pizza was delivered cold and the cheese was n \' t even fully melted !"

\begin{itemize}[topsep=0pt,itemsep=-1ex,partopsep=1ex,parsep=1ex]
\item Category term: "food quality"
\item Aspect term: "cheese"
\item Sentiment term: "negative"
\item Target Opinion term: "was n \' t even fully melted"
\end{itemize}

\textit{\#\# Test Sample}

Input sentence:

        \end{tcolorbox}
    \end{subtable}

\begin{subtable}{\linewidth}
\begin{tcolorbox}[
    colback=orangebg,
    colframe=black,
    fontupper=\footnotesize
]

\textit{\#\# System Prompt}

Given an input sentence, Sentiment, Opinion, Categroy, and \textbf{target Aspect terms}, extract various expressions that extend the span of the aspect term. The new expressions should be formed by adding surrounding words and must meet the following criteria:

\begin{enumerate}[topsep=0pt,itemsep=-1ex,partopsep=1ex,parsep=1ex]
    \item Be related to the given aspect term.
    \item Include neighboring words around the aspect term to form a new expression.
    \item Should not overlap with the original opinion term [O].
\item While expanding the aspect term, avoid incorporating the entire sentence.
\item If it is not feasible to expand the expression, return only the original aspect term.
\item Ensure the new expression matches exactly as it appears in the input sentence.
\end{enumerate}

\textit{\#\# Demonstration}

Input sentence: "quacamole at pacifico is yummy , as are the wings with chimmichuri ."
\begin{itemize}[topsep=0pt,itemsep=-1ex,partopsep=1ex,parsep=1ex]
\item Sentiment term: "positive"
\item Opinion term: "yummy"
\item Category term: "food quality"
\item Target Aspect term: "quacamole"
\end{itemize}

\textit{\#\# Test Sample}

Input sentence:

\end{tcolorbox}

\end{subtable}
\caption{Full prompt for \opinion{Opinion} terms in \narrow and \aspect{Aspect} terms in \widen, respectively.}
\label{tab:icl_prompt}
\end{table*}


\begin{table*}[ht]
\centering
\resizebox{.9\textwidth}{!}{%
\begin{tcolorbox}[
    colback=promptgray,
    colframe=black,
    fontupper=\footnotesize
]
Aspect-based sentiment analysis aims to identify the aspects of given target entities and the sentiment expressed towards each aspect.

For example, from an example sentence: ``This restaurant is rude, but the food is delicious'',
we can extract the negative sentiment that the restaurant is (1) ``rude'' in terms of ``service general'' and (2) ``delicious'' in terms of ``food quality''.
As such, the complex task of categorizing the aspect terms and their corresponding categories and the sentiment expressed for the aspect in the sentence into one of three classes [positive, negative, neutral] is the Aspect-based sentiment analysis (ABSA) task.

Each element that is extracted is called an element, and the characteristics of each element can be described as follows.

\begin{enumerate}
\vspace{-2mm}
\item Aspect: The aspect covered by the sentence, such as restaurant, food name, or service.
\begin{itemize}
\vspace{-1mm}
\item[-] Any phrase, verb, or noun that mentions a particular aspect can be an aspect.
\item[-] Aspects can be extracted with or without quotation marks.
\item[-] Determiners are excluded unless they are part of a noun phrase.
\item[-] Subjectivity indicators that indicate opinion are not included.
\item[-] Specific product names are not aspect terms.
\item[-] Even if pronouns refer to an aspect, they are not aspect terms.
\item[-] Pronouns (e.g., ``it'', ``they'', ``this'') even if they refer to an aspect.
\item[-] If they appear in the sentence, we extract their span as an aspect; if they do not appear directly in the sentence, we define `null' as the aspect term.
\end{itemize}
\vspace{-2mm}
\item Category: Predefined categories to categorize aspects. Categories are divided into two levels: six entity categories and five attribute labels, each characterized by the following features:
\begin{itemize}
    \item[1)] Entity
    \begin{itemize}
        \item[a.] FOOD for opinions focusing on the food in general or in terms of specific dishes, dining options etc.
        \item[b.] DRINKS for opinions focusing on the drinks in general or in terms of specific drinks, drinking options etc.
        \item[c.] SERVICE for opinions focusing on the (customer/kitchen/counter) service, on the promptness and quality of the restaurant's service in general, the food preparation, the staff's attitude and professionalism, the wait time, the options offered (e.g. takeout), etc
        \item[d.] AMBIENCE for opinions focusing on the atmosphere or the environment of the restaurant's interior or exterior space (e.g. terrace, yard, garden), the décor, entertainment options, etc.
        \item[e.] LOCATION for opinions focusing on the location of the reviewed restaurant in terms of its position, the surroundings, the view, etc.
        \item[f.] RESTAURANT for opinions expressed about the (specific) evaluated restaurant as whole not focusing on any of the above five entity types.
    \end{itemize}

    \item[2)] Attribute
    \begin{itemize}
        \item[a.] GENERAL. This attribute label is assigned to sentences that express general positive or negative sentiment about an entity type.
        \item[b.] PRICES for opinions that refer to the prices of the food, the drinks or the restaurant in general. e.g.
        \item[c.] QUALITY for opinions focusing on the taste, the freshness, the texture, the consistency, the temperature, the preparation, the authenticity, the cooking or general quality of the food and the drinks served in the restaurant.
        \item[d]. STYLE\&OPTIONS for opinions referring to the presentation, the serving style, the portions size, the food/menu options or variety (e.g. innovative dishes/drinks, vegetarian options) of the food and of the drinks served in the restaurant.
        \item[e.] MISCELLANEOUS for attributes that do not fall into any of the aforementioned cases.
    \end{itemize}
    \item[3)] Final Entity Attribute pairs
We end up with 13 Entity Attribute category pairs: 
['location general', `food prices', `food quality', `food general', `food style\&options', `ambience general', `service general', `restaurant general', `restaurant prices', `restaurant miscellaneous', `drinks prices', `drinks quality', `drinks style\&options'].
\end{itemize}
\vspace{-2mm}
\item Opinion: An opinion term that expresses a sentiment about an aspect. If it appears directly in the sentence as a single word or phrase, we extract it. However, if no specific phrase can be extracted, and the sentiment about the aspect can be gleaned from the nuances of the sentence as a whole, we define `null' as the opinion term.
\vspace{-2mm}
\item Sentiment: The sentiment expressed by the customer about an aspect, divided into three classes: [Positive, Neutral, Negative]. The neutral label applies for mildly positive or negative sentiment, thus it does not indicate objectivity
\end{enumerate}

\vspace{-1mm}

To summarize, we want to extract one or more quadruples of (aspect, category, opinion, sentiment) from a given review. As mentioned before, the aspect term and opinion term can be extracted as `null' if they are not evident in the sentence, while category and sentiment must be selected from the predefined classes ``2. Category'' and ``4. Sentiment''.

It is up to the model to decide in which order to predict each element of the quadruple. The model is given 20 examples, as shown below, and the sentence you want to test.

Place the aspect term after [A], the opinion term after [O], the sentiment polarity after [S], and the aspect category after [C] in the format `ORDER'. If multiple quadruples are predicted, insert `\#\#\#\#' to separate the quadruples.
\end{tcolorbox}%
}
\caption{System prompt of ACOS-Rest datasets, which LLMs utilize to make the prediction, in Section~\ref{sec:model-performance}. `ORDER' is a hyperparameter that guides the model in a specific element order. The variants of score according to order are described in Table~\ref{tab:acos-asqp-gpt}, and Table~\ref{tab:acos-asqp-gemini}.}
\label{tab:model_pred_prompt}
\end{table*}

\begin{table*}[ht]
        \begin{tcolorbox}[
            colback=promptgray,
            colframe=black,
            fontupper=\footnotesize
        ]
        \textbf{User:}
        \begin{itemize}[topsep=0pt,itemsep=-1ex,partopsep=1ex,parsep=1ex]
            \item Input Sentence: the pizza was delivered cold and the cheese was n' t even fully melted ! 
        
    \item GT: [`cheese', `food quality', `negative', `n' t even fully melted'],
  
  \item New Opinion Term: even fully melted
  \end{itemize}

\textbf{AI:}

\quad The input sentence details a quality issue with the pizza, explicitly noting that the [A] ``cheese'' was not adequately melted, articulated through the opinion [O] ``n ' t even fully melted.' The new opinion term under evaluation is ``even fully melted.'' Let's consider the criteria step by step:

\begin{enumerate}[topsep=0pt,itemsep=-1ex,partopsep=1ex,parsep=1ex]
    \item Aspect and Category Relevance:
    \begin{itemize}[topsep=0pt,itemsep=-1ex,partopsep=1ex,parsep=1ex]
        \item[-] Reasoning: The new opinion term ``even fully melted'' is derived from the original opinion term [O] ``n ' t even fully melted'' related to [A] ``cheese'' and pertains directly to [C] ``food quality,'' focusing specifically on the melting quality of the cheese. The new term aligns with the original aspect and category.
        \item[-] Decision: True
    \end{itemize}

 \item Sentiment and Opinion Consistency:
\begin{itemize}[topsep=0pt,itemsep=-1ex,partopsep=1ex,parsep=1ex]
    \item[-] Reasoning: The new term ``even fully melted'' captures part of the original opinion term [O] ``n \' t even fully melted.'' However, it omits the critical negation ``n ' t,'' which is essential to convey the negative sentiment [S] ``negative'' associated with the GT. Without this negation, the sentiment changes to positive. Thus, the new term does not maintain the sentiment polarity.
    \item[-] Decision: False
\end{itemize}

 \item Extractability:
    \begin{itemize}[topsep=0pt,itemsep=-1ex,partopsep=1ex,parsep=1ex]
        \item[-] Reasoning: ``Even fully melted'' is a phrase directly extracted from the end of the sentence without adding or significantly altering the words.
        \item[-] Decision: True
\end{itemize}
 \item Independence:
 \begin{itemize}[topsep=0pt,itemsep=-1ex,partopsep=1ex,parsep=1ex]
     \item[-] Reasoning: The new opinion term is ``even fully melted,'' and the aspect term is [A] ``cheese.'' The new term and aspect term are independent, with no overlap.
    \item[-] Decision: True
\end{itemize}
\end{enumerate}

Judgment: Since the new term ``even fully melted'' fails to meet the Sentiment and Opinion Consistency criterion, it is deemed invalid.
]
        \end{tcolorbox}
    \caption{Demonstration example of \judge. }
\label{tab:llm_demo_example}
\end{table*}
\begin{table*}[ht]
    
\begin{subtable}{\linewidth}
\begin{tcolorbox}[
    colback=orangebg,
    colframe=black,
    fontupper=\footnotesize
]
\textbf{User:}
\begin{itemize}[topsep=0pt,itemsep=-1ex,partopsep=1ex,parsep=1ex]
    \item Input sentence: great place to relax and enjoy your dinner
    \item GT: [A] place [C] ambience general [S] positive [O] great
    \item New Aspect term: great place
 \end{itemize}

 \textbf{AI:}
 
 \quad The input sentence describes a ``great place to relax and enjoy your dinner.'' The new aspect term under evaluation is ``great place''. Let's consider the criteria step by step:
 \begin{enumerate}[topsep=0pt,itemsep=-1ex,partopsep=1ex,parsep=1ex]
     \item Aspect and Category Consistency:
     \begin{itemize}[topsep=0pt,itemsep=-1ex,partopsep=1ex,parsep=1ex]
         \item[-] Reasoning: The original GT aspect term [A] "place" falls under the category [C] "ambience general." The new term "great place" still refers to the same aspect of the place, maintaining the focus on the ambiance. The new term aligns with the original aspect and category.
         \item[-] Decision: True
     \end{itemize}
    \item Sentiment and Opinion Relevance:
    \begin{itemize}[topsep=0pt,itemsep=-1ex,partopsep=1ex,parsep=1ex]
        \item[-] Reasoning: The original opinion term [O] ``great'' describes the positive sentiment towards the ``place.'' The new term ``great place'' combines the aspect and opinion into one phrase, which is not allowed as it should be independent. The sentiment and opinion relevance are not maintained correctly.
        \item[-] Decision: False
    \end{itemize}
    \item Extractability
    \begin{itemize}[topsep=0pt,itemsep=-1ex,partopsep=1ex,parsep=1ex]
        \item[-] Reasoning: The new term ``great place'' is directly taken from the sentence without adding new words or significantly rearranging existing ones. However, it combines the aspect and opinion into one phrase, which is not allowed.
        \item[-] Decision: False
    \end{itemize}
    \item Independence
        \begin{itemize}[topsep=0pt,itemsep=-1ex,partopsep=1ex,parsep=1ex]
            \item[-] Reasoning: The new aspect term ``great place'' incorporates the opinion term [O] ``great,'' which violates the independence criterion. The aspect term should not include the opinion term.
            \item[-] Decision: False
        \end{itemize}
 \end{enumerate}
 
\quad Judgment: Since the new term ``great place'' fails to meet the criteria for Sentiment and Opinion Relevance, Extractability, and Independence, it is deemed invalid.
\end{tcolorbox}

\end{subtable}
\begin{subtable}{\linewidth}
        \begin{tcolorbox}[
            colback=bluebg,
            colframe=black,
            fontupper=\footnotesize
        ]

        \textbf{User:}
\begin{itemize}[topsep=0pt,itemsep=-1ex,partopsep=1ex,parsep=1ex]
    \item Input sentence: i ordered one , the touch pad failed to work consistently .
    \item GT: [A] touch pad [C] hardware [S] negative [O] failed
    \item New Opinon Term: failed to work
 \end{itemize}

\textbf{AI:}
 
 \quad The input sentence mentions that the touch pad failed to work consistently. The new opinion term under evaluation is ``failed to work''. Let's consider the criteria step by step:
 \begin{enumerate}[topsep=0pt,itemsep=-1ex,partopsep=1ex,parsep=1ex]
     \item Aspect and Category Consistency:
     \begin{itemize}[topsep=0pt,itemsep=-1ex,partopsep=1ex,parsep=1ex]
         \item[-] Reasoning: The original GT opinion term is [O] ``failed,'' which describes the [A] ``touch pad'' in the [C] ``hardware'' category. The new term "failed to work" still describes the touch pad's functionality, aligning with the original aspect and category.
         \item[-] Decision: True
     \end{itemize}
    \item Sentiment and Opinion Relevance:
    \begin{itemize}[topsep=0pt,itemsep=-1ex,partopsep=1ex,parsep=1ex]
        \item[-] Reasoning: The original GT opinion term is [O] ``failed,'' which conveys a negative sentiment. The new term ``failed to work'' maintains the same negative sentiment and expands on the original opinion by specifying the failure to work. The sentiment and opinion consistency are preserved.
        \item[-] Decision: True
    \end{itemize}
    \item Extractability
    \begin{itemize}[topsep=0pt,itemsep=-1ex,partopsep=1ex,parsep=1ex]
        \item[-] Reasoning: The new term ``failed to work'' can be directly extracted from the sentence without adding new words or significantly rearranging existing ones.
        \item[-] Decision: True
    \end{itemize}
    \item Independence
        \begin{itemize}[topsep=0pt,itemsep=-1ex,partopsep=1ex,parsep=1ex]
            \item[-] Reasoning: The new opinion term ``failed to work'' and the GT aspect term [A] ``touch pad'' are independent of each other. The opinion term does not include any part of the aspect term, ensuring no overlap.
            \item[-] Decision: True
        \end{itemize}
 \end{enumerate}
 
\quad Judgment: The new opinion term ``failed to work'' meets all the criteria and is deemed valid.
\end{tcolorbox}
    \end{subtable}

\caption{Output example of \judge: invalid \aspect{aspect} and valid \opinion{opinion} terms.}
\label{tab:judge_ouptut}
\end{table*}


\begin{table*}[t]
\centering
\resizebox{\textwidth}{!}{%
\begin{tcolorbox}[
    colback=promptgray,
    colframe=black,
    fontupper=\footnotesize
]
\begin{enumerate}[topsep=-0.5pt,itemsep=0ex,partopsep=.8ex,parsep=.8ex]
    \item \textbf{\textbf{User:}} although the tables may be closely situated , the candle - light , food - quality and service overcompensate .
    \vspace{-1mm}
    
    \textbf{AI:} [A] tables [C] ambience general [S] negative [O] closely situated \#\#\#\# [A] candle - light [C] ambience general [S] positive [O] overcompensate \#\#\#\# [A] food [C] food quality [S] positive [O] overcompensate \#\#\#\# [A] service [C] service general [S] positive [O] overcompensate
    
    \item \textbf{User:} these are overpriced and you can get better just around the corner :
    \vspace{-1mm}
    
    \textbf{AI:} [A] null [C] food prices [S] negative [O] overpriced \#\#\#\# [A] null [C] food quality [S] negative [O] null

\item \textbf{User:} they have authentic indian at amazing prices .

    \vspace{-1mm}
    \textbf{AI:} [A] indian [C] food quality [S] positive [O] authentic \#\#\#\# [A] null [C] food prices [S] positive [O] amazing

\item \textbf{User:} prices too high for this cramped and unappealing resturant .
    \vspace{-1mm}
    
    \textbf{AI:} [A] resturant [C] restaurant prices [S] negative [O] high \#\#\#\# [A] resturant [C] ambience general [S] negative [O] cramped \#\#\#\# [A] resturant [C] ambience general [S] negative [O] unappealing

\item \textbf{User:} do n ' t get me started on the margaritas , either .
    \vspace{-1mm}
    
    \textbf{AI:} [A] margaritas [C] drinks quality [S] negative [O] null

\item \textbf{User:} \$ 6 and there is much tasty food , all of it fresh and continually refilled .
\vspace{-1mm}

\textbf{AI:} [A] food [C] food style\&options [S] positive [O] refilled \#\#\#\# [A] food [C] food quality [S] positive [O] tasty \#\#\#\# [A] food [C] food quality [S] positive [O] fresh \#\#\#\# [A] food [C] food prices [S] positive [O] null

\item \textbf{User:} located at the end of a magnificent block .

\vspace{-1mm}
\textbf{AI:} [A] null [C] location general [S] positive [O] magnificent

\item \textbf{User:} the staff has been nice , but they seemed really stressed and the unisex bathroom needs to be cleaned more often .

    \vspace{-1mm}
    \textbf{AI:} [A] staff [C] service general [S] positive [O] nice \#\#\#\# [A] staff [C] service general [S] positive [O] stressed \#\#\#\# [A] unisex bathroom [C] ambience general [S] negative [O] null

\item \textbf{User:} and amazingly cheap .
\vspace{-1mm}

    \textbf{AI:} [A] null [C] food prices [S] positive [O] amazingly

\item \textbf{User:} cheese plate is a varied delight and great bargain at \$ 10 .
\vspace{-1mm}

    \textbf{AI:} [A] cheese plate [C] food quality [S] positive [O] varied delight \#\#\#\# [A] cheese plate [C] food style\&options [S] positive [O] varied delight \#\#\#\# [A] cheese plate [C] food prices [S] positive [O] great bargain

\item   \textbf{User:} not what i would expect for the price and prestige of this location .
\vspace{-1mm}

    \textbf{AI:} [A] location [C] restaurant prices [S] neutral [O] null \#\#\#\# [A] location [C] restaurant miscellaneous [S] neutral [O] null \#\#\#\# [A] null [C] service general [S] negative [O] null

\item \textbf{User:} terrible waste of money . . scammers
\vspace{-1mm}

    \textbf{AI:} [A] null [C] restaurant general [S] negative [O] scammers \#\#\#\# [A] null [C] restaurant prices [S] negative [O] terrible

\item \textbf{User:} small servings for main entree , i had salmon ( wasnt impressed ) girlfriend had chicken , it was good .
\vspace{-1mm}
    
    \textbf{AI:} [A] salmon [C] food quality [S] negative [O] wasnt impressed \#\#\#\# [A] chicken [C] food quality [S] positive [O] good \#\#\#\# [A] servings for main entree [C] food general [S] negative [O] small

\item \textbf{User:} first it took us a long time to find the place .
    \vspace{-1mm}
    
    \textbf{AI:} [A] place [C] restaurant miscellaneous [S] negative [O] null

\item \textbf{User:} they were such a rip - off ( \$ 8 . 95 for four small meat patties in steamed buns ) and not worth trying .
\vspace{-1mm}

    \textbf{AI:} [A] null [C] food quality [S] negative [O] not worth \#\#\#\# [A] null [C] food style\&options [S] negative [O] small \#\#\#\# [A] null [C] food prices [S] negative [O] rip - off

\item \textbf{User:} restaurant with a view
\vspace{-1mm}

    \textbf{AI:} [A] view [C] location general [S] neutral [O] null

\item \textbf{User:} the lobster sandwich is \$ 24 and although it was good it was not nearly enough to warrant that price .
\vspace{-1mm}

    \textbf{AI:} [A] lobster sandwich [C] food quality [S] positive [O] good \#\#\#\# [A] lobster sandwich [C] food prices [S] negative [O] not nearly enough

\item \textbf{User:} my husbands birthday and my sons was not as it was intended . . . and we drove two hours to spend too much money to be treated terribly !
    \vspace{-1mm}
    
    \textbf{AI:} [A] null [C] restaurant general [S] negative [O] null \#\#\#\# [A] null [C] restaurant prices [S] negative [O] null \#\#\#\# [A] null [C] service general [S] negative [O] terribly

\item \textbf{User:} the bar is very well stocked with interesting beers and well priced wines .
\vspace{-1mm}

    \textbf{AI:} [A] bar [C] drinks style\&options [S] positive [O] well stocked \#\#\#\# [A] beers [C] drinks style\&options [S] positive [O] interesting \#\#\#\# [A] wines [C] drinks prices [S] positive [O] well priced

\item \textbf{User:} the decor however seems to be the distraction so you wo n ' t notice that you just payed 300 bucks for some cold eggplant that took 2 frickin hours to come ! ! ! !
\vspace{-1mm}

    \textbf{AI:} [A] decor [C] ambience general [S] neutral [O] distraction \#\#\#\# [A] eggplant [C] food quality [S] negative [O] cold \#\#\#\# [A] eggplant [C] food prices [S] negative [O] null \#\#\#\# [A] null [C] service general [S] negative [O] null
\end{enumerate}
\end{tcolorbox}%
\vspace{-3mm}
}
\caption{20-shot examples of ACOS-Rest dataset utilized in Section~\ref{sec:model-performance}. ``\#\#\#\#'' is a separate token that distinguish the quadruples.}
\label{tab:demonstration_examples}
\end{table*}

\begin{table*}[t]
\centering
\resizebox{\textwidth}{!}{%
\begin{tabular}{c|ccc|ccc|ccc|ccc}
\toprule
\multirow{2}{*}{\textbf{\textit{GPT-3.5}}} & \multicolumn{6}{c|}{\textbf{ASQP}} & \multicolumn{6}{c}{\textbf{ACOS}} \\  
& \multicolumn{3}{c|}{\textbf{Rest15}} & \multicolumn{3}{c|}{\textbf{Rest16}} & \multicolumn{3}{c|}{\textbf{Laptop}} & \multicolumn{3}{c}{\textbf{Rest}}  \\ 
\midrule
\textbf{Orders} & \textbf{Origin} & \textbf{Ours} & \textbf{$\Delta$} & \textbf{Origin} & \textbf{Ours} & \textbf{$\Delta$} & \textbf{Origin} & \textbf{Ours} & \textbf{$\Delta$} & \textbf{Origin} & \textbf{Ours} & \textbf{$\Delta$} \\
\midrule
AOSC & 28.85 & 42.87 & \cellcolor{maincolor!55} 14.02 & 37.81 & 50.53 & \cellcolor{maincolor!57} 12.72 & 25.66 & 30.5 & \cellcolor{maincolor!35} 4.84 & 42.34 & 50.18 & \cellcolor{maincolor!38} 7.84 \\
OCSA & 32.44 & 43.88 & \cellcolor{maincolor!15} 11.44 & 31.61 & 41.17 & \cellcolor{maincolor!0} 9.56 & 27.35 & 31.18 & \cellcolor{maincolor!17} 3.83 & 36.58 & 44.31 & \cellcolor{maincolor!34} 7.73 \\
OSAC & 32.89 & 43.7 & \cellcolor{maincolor!6} 10.81 & 35.97 & 48.18 & \cellcolor{maincolor!48} 12.21 & 28.18 & 32.19 & \cellcolor{maincolor!20} 4.01 & 41.32 & 49.21 & \cellcolor{maincolor!40} 7.89 \\
OSCA & 32.09 & 44.01 & \cellcolor{maincolor!23} 11.92 & 34.62 & 47.25 & \cellcolor{maincolor!55} 12.63 & 25.17 & 28.03 & \cellcolor{maincolor!0} 2.86 & 34.61 & 41.46 & \cellcolor{maincolor!0} 6.85 \\
OACS & 32.27 & 44.11 & \cellcolor{maincolor!21} 11.84 & 36.61 & 48.45 & \cellcolor{maincolor!41} 11.84 & 29.06 & 32.91 & \cellcolor{maincolor!17} 3.85 & 39.96 & 49.0 & \cellcolor{maincolor!84} 9.04 \\
AOCS & 29.95 & 45.5 & \cellcolor{maincolor!78} 15.55 & 35.36 & 48.62 & \cellcolor{maincolor!67} 13.26 & 23.39 & 28.55 & \cellcolor{maincolor!41} 5.16 & 37.86 & 45.78 & \cellcolor{maincolor!41} 7.92 \\
COAS & 32.48 & 44.48 & \cellcolor{maincolor!24} 12.0 & 35.31 & 46.48 & \cellcolor{maincolor!29} 11.17 & 24.94 & 28.91 & \cellcolor{maincolor!19} 3.97 & 35.95 & 44.96 & \cellcolor{maincolor!83} 9.01 \\
SAOC & 29.1 & 43.59 & \cellcolor{maincolor!62} 14.49 & 36.54 & 50.31 & \cellcolor{maincolor!76} 13.77 & 25.22 & 31.22 & \cellcolor{maincolor!56} 6.0 & 40.78 & 49.5 & \cellcolor{maincolor!72} 8.72 \\
OASC & \textbf{34.33} & 44.72 & \cellcolor{maincolor!0} 10.39 & \textbf{39.64} & \textbf{51.22} & \cellcolor{maincolor!36} 11.58 & 26.69 & 30.52 & \cellcolor{maincolor!17} 3.83 & \textbf{43.4} & \textbf{51.74} & \cellcolor{maincolor!57} 8.34 \\
SOAC & 30.49 & 42.46 & \cellcolor{maincolor!23} 11.97 & 36.26 & 48.57 & \cellcolor{maincolor!50} 12.31 & 23.7 & 27.58 & \cellcolor{maincolor!18} 3.88 & 39.62 & 48.16 & \cellcolor{maincolor!65} 8.54 \\
SOCA & 31.19 & 44.52 & \cellcolor{maincolor!44} 13.33 & 30.98 & 40.95 & \cellcolor{maincolor!7} 9.97 & \textbf{40.13} & \textbf{48.57} & \cellcolor{maincolor!100} \textbf{8.44} & 38.14 & 45.67 & \cellcolor{maincolor!26} 7.53 \\
ASOC & 28.81 & 43.68 & \cellcolor{maincolor!67} 14.87 & 36.79 & 50.11 & \cellcolor{maincolor!68} 13.32 & 24.41 & 29.4 & \cellcolor{maincolor!38} 4.99 & 38.43 & 46.71 & \cellcolor{maincolor!55} 8.28 \\
CAOS & 28.23 & 44.09 & \cellcolor{maincolor!82} 15.86 & 36.91 & 49.66 & \cellcolor{maincolor!58} 12.75 & 22.89 & 29.3 & \cellcolor{maincolor!63} 6.41 & 36.94 & 45.32 & \cellcolor{maincolor!59} 8.38 \\
SCAO & 26.55 & 43.03 & \cellcolor{maincolor!92} 16.48 & 34.41 & 49.26 & \cellcolor{maincolor!96} 14.85 & 26.86 & 33.37 & \cellcolor{maincolor!65} 6.51 & 35.09 & 43.91 & \cellcolor{maincolor!76} 8.82 \\
OCAS & 33.95 & \textbf{46.83} & \cellcolor{maincolor!37} 12.88 & 36.14 & 47.43 & \cellcolor{maincolor!31} 11.29 & 30.61 & 34.01 & \cellcolor{maincolor!9} 3.4 & 39.83 & 48.78 & \cellcolor{maincolor!81} 8.95 \\
COSA & 27.85 & 40.04 & \cellcolor{maincolor!27} 12.19 & 33.68 & 44.25 & \cellcolor{maincolor!18} 10.57 & 24.62 & 27.93 & \cellcolor{maincolor!8} 3.31 & 29.69 & 36.79 & \cellcolor{maincolor!9} 7.1 \\
CASO & 26.92 & 41.05 & \cellcolor{maincolor!56} 14.13 & 33.65 & 48.7 & \cellcolor{maincolor!100} \textbf{15.05} & 24.12 & 28.9 & \cellcolor{maincolor!34} 4.78 & 33.89 & 41.86 & \cellcolor{maincolor!43} 7.97 \\
CSAO & 26.34 & 43.33 & \cellcolor{maincolor!100} \textbf{16.99} & 33.59 & 48.42 & \cellcolor{maincolor!95} 14.83 & 25.29 & 31.75 & \cellcolor{maincolor!64} 6.46 & 36.33 & 45.76 & \cellcolor{maincolor!100} \textbf{9.43} \\
ACOS & 26.06 & 41.74 & \cellcolor{maincolor!80} 15.68 & 33.32 & 46.4 & \cellcolor{maincolor!64} 13.08 & 24.39 & 30.73 & \cellcolor{maincolor!62} 6.34 & 35.88 & 44.14 & \cellcolor{maincolor!54} 8.26 \\
ACSO & 25.79 & 40.7 & \cellcolor{maincolor!68} 14.91 & 32.03 & 46.24 & \cellcolor{maincolor!84} 14.21 & 23.19 & 27.89 & \cellcolor{maincolor!32} 4.7 & 31.85 & 41.25 & \cellcolor{maincolor!98} 9.4 \\
SCOA & 28.95 & 42.2 & \cellcolor{maincolor!43} 13.25 & 33.7 & 46.15 & \cellcolor{maincolor!52} 12.45 & 23.25 & 27.59 & \cellcolor{maincolor!26} 4.34 & 34.28 & 42.93 & \cellcolor{maincolor!69} 8.65 \\
CSOA & 28.84 & 40.97 & \cellcolor{maincolor!26} 12.13 & 34.21 & 46.73 & \cellcolor{maincolor!53} 12.52 & 23.2 & 27.79 & \cellcolor{maincolor!31} 4.59 & 34.22 & 42.95 & \cellcolor{maincolor!72} 8.73 \\
SACO & 26.49 & 41.74 & \cellcolor{maincolor!73} 15.25 & 34.29 & 47.75 & \cellcolor{maincolor!71} 13.46 & 26.43 & 31.76 & \cellcolor{maincolor!44} 5.33 & 33.63 & 42.14 & \cellcolor{maincolor!64} 8.51 \\
ASCO & 25.91 & 41.71 & \cellcolor{maincolor!81} 15.8 & 32.84 & 47.61 & \cellcolor{maincolor!94} 14.77 & 23.31 & 28.76 & \cellcolor{maincolor!46} 5.45 & 31.99 & 39.94 & \cellcolor{maincolor!42} 7.95 \\
\midrule
\textbf{Avg.} & 29.45 & 43.12 & 13.67 & 34.84 & 47.52 & 12.67 & 25.92 & 30.81 & 4.89 & 36.78 & 45.1 & 8.33 \\
\textbf{Std.} & 2.71 & 1.63 & 1.91 & 2.06 & 2.57 & 1.51 & 3.64 & 4.25 & 1.31 & 3.5 & 3.63 & 0.66 \\

\bottomrule
\end{tabular}%
}
\caption{F1 scores of GPT-3.5-Turbo. The \textbf{best score} for each column is bold.}
\label{tab:acos-asqp-gpt}
\end{table*}

\begin{table*}[t]
\centering
\resizebox{\textwidth}{!}{%
\begin{tabular}{c|ccc|ccc|ccc|ccc}
\toprule
\multirow{2}{*}{\textbf{\textit{GPT-4o-mini}}} & \multicolumn{6}{c|}{\textbf{ASQP}} & \multicolumn{6}{c}{\textbf{ACOS}} \\  
& \multicolumn{3}{c|}{\textbf{Rest15}} & \multicolumn{3}{c|}{\textbf{Rest16}} & \multicolumn{3}{c|}{\textbf{Laptop}} & \multicolumn{3}{c}{\textbf{Rest}}  \\ 
\midrule
\textbf{Orders} & \textbf{Origin} & \textbf{Ours} & \textbf{$\Delta$} & \textbf{Origin} & \textbf{Ours} & \textbf{$\Delta$} & \textbf{Origin} & \textbf{Ours} & \textbf{$\Delta$} & \textbf{Origin} & \textbf{Ours} & \textbf{$\Delta$} \\
\midrule
AOSC & 33.28 & 48.91 & \cellcolor{maincolor!67} 15.63 & 42.54 & 54.59 & \cellcolor{maincolor!25} 12.05 & 32.3 & 37.45 & \cellcolor{maincolor!38} 5.15 & 37.09 & 44.72 & \cellcolor{maincolor!28} 7.63 \\
OCSA & \textbf{38.45} & 49.89 & \cellcolor{maincolor!0} 11.44 & 43.41 & 54.31 & \cellcolor{maincolor!5} 10.9 & 34.98 & 39.42 & \cellcolor{maincolor!17} 4.44 & 42.16 & 50.08 & \cellcolor{maincolor!38} 7.92 \\
OSAC & 37.25 & 49.86 & \cellcolor{maincolor!18} 12.61 & 43.7 & 54.68 & \cellcolor{maincolor!6} 10.98 & 35.39 & 39.93 & \cellcolor{maincolor!20} 4.54 & 41.99 & 49.69 & \cellcolor{maincolor!30} 7.7 \\
OSCA & 36.12 & 48.01 & \cellcolor{maincolor!7} 11.89 & 42.86 & 53.46 & \cellcolor{maincolor!0} 10.6 & 34.84 & 39.69 & \cellcolor{maincolor!29} 4.85 & 41.08 & 48.7 & \cellcolor{maincolor!27} 7.62 \\
OACS & 36.9 & 48.71 & \cellcolor{maincolor!5} 11.81 & 41.46 & 53.3 & \cellcolor{maincolor!21} 11.84 & \textbf{37.38} & \textbf{41.97} & \cellcolor{maincolor!21} 4.59 & 40.23 & 47.67 & \cellcolor{maincolor!21} 7.44 \\
AOCS & 32.7 & 49.38 & \cellcolor{maincolor!84} 16.68 & 42.55 & 56.06 & \cellcolor{maincolor!51} 13.51 & 32.69 & 38.74 & \cellcolor{maincolor!64} 6.05 & 39.23 & 46.39 & \cellcolor{maincolor!11} 7.16 \\
COAS & 37.1 & 50.68 & \cellcolor{maincolor!34} 13.58 & 43.26 & 56.51 & \cellcolor{maincolor!46} 13.25 & 33.43 & 39.23 & \cellcolor{maincolor!56} 5.8 & 41.1 & 49.32 & \cellcolor{maincolor!49} 8.22 \\
SAOC & 32.18 & 49.01 & \cellcolor{maincolor!87} 16.83 & 41.05 & 54.7 & \cellcolor{maincolor!53} 13.65 & 32.65 & 37.89 & \cellcolor{maincolor!40} 5.24 & 39.14 & 46.91 & \cellcolor{maincolor!33} 7.77 \\
OASC & 37.23 & 49.29 & \cellcolor{maincolor!10} 12.06 & 42.14 & 53.73 & \cellcolor{maincolor!17} 11.59 & 34.86 & 38.69 & \cellcolor{maincolor!0} 3.83 & 41.77 & 48.89 & \cellcolor{maincolor!9} 7.12 \\
SOAC & 35.82 & 48.7 & \cellcolor{maincolor!23} 12.88 & 41.02 & 53.22 & \cellcolor{maincolor!28} 12.2 & 34.75 & 39.77 & \cellcolor{maincolor!34} 5.02 & 40.27 & 47.22 & \cellcolor{maincolor!3} 6.95 \\
SOCA & 36.82 & 48.99 & \cellcolor{maincolor!11} 12.17 & 42.63 & 53.68 & \cellcolor{maincolor!7} 11.05 & 34.48 & 39.97 & \cellcolor{maincolor!47} 5.49 & 38.87 & 45.75 & \cellcolor{maincolor!1} 6.88 \\
ASOC & 31.5 & 47.59 & \cellcolor{maincolor!75} 16.09 & 41.32 & 53.99 & \cellcolor{maincolor!36} 12.67 & 31.51 & 36.3 & \cellcolor{maincolor!27} 4.79 & 38.41 & 45.26 & \cellcolor{maincolor!0} 6.85 \\
CAOS & 33.12 & 50.09 & \cellcolor{maincolor!89} 16.97 & 43.08 & \textbf{57.32} & \cellcolor{maincolor!63} 14.24 & 34.79 & 41.03 & \cellcolor{maincolor!69} 6.24 & 39.96 & 48.76 & \cellcolor{maincolor!70} 8.8 \\
SCAO & 32.17 & 49.8 & \cellcolor{maincolor!100} \textbf{17.63} & 40.81 & 56.48 & \cellcolor{maincolor!88} 15.67 & 31.87 & 39.16 & \cellcolor{maincolor!100} \textbf{7.29} & 38.66 & 48.28 & \cellcolor{maincolor!100} \textbf{9.62} \\
OCAS & 37.94 & 49.69 & \cellcolor{maincolor!5} 11.75 & \textbf{44.36} & 55.36 & \cellcolor{maincolor!7} 11.0 & 36.32 & 41.42 & \cellcolor{maincolor!36} 5.1 & 40.52 & 47.42 & \cellcolor{maincolor!1} 6.9 \\
COSA & 37.39 & \textbf{51.01} & \cellcolor{maincolor!35} 13.62 & 44.11 & 55.78 & \cellcolor{maincolor!18} 11.67 & 33.24 & 38.75 & \cellcolor{maincolor!48} 5.51 & 42.67 & \textbf{51.0} & \cellcolor{maincolor!53} 8.33 \\
CASO & 31.15 & 48.49 & \cellcolor{maincolor!95} 17.34 & 41.73 & 56.69 & \cellcolor{maincolor!76} 14.96 & 33.4 & 39.72 & \cellcolor{maincolor!71} 6.32 & 38.2 & 46.05 & \cellcolor{maincolor!36} 7.85 \\
CSAO & 31.1 & 48.71 & \cellcolor{maincolor!99} 17.61 & 39.87 & 56.17 & \cellcolor{maincolor!100} \textbf{16.3} & 32.45 & 39.65 & \cellcolor{maincolor!97} 7.2 & 41.59 & 50.99 & \cellcolor{maincolor!92} 9.4 \\
ACOS & 28.92 & 43.95 & \cellcolor{maincolor!57} 15.03 & 40.51 & 53.64 & \cellcolor{maincolor!44} 13.13 & 34.64 & 40.36 & \cellcolor{maincolor!54} 5.72 & 36.25 & 45.21 & \cellcolor{maincolor!76} 8.96 \\
ACSO & 29.22 & 45.37 & \cellcolor{maincolor!76} 16.15 & 39.51 & 51.94 & \cellcolor{maincolor!32} 12.43 & 32.39 & 38.07 & \cellcolor{maincolor!53} 5.68 & 36.84 & 45.09 & \cellcolor{maincolor!50} 8.25 \\
SCOA & 35.61 & 49.52 & \cellcolor{maincolor!39} 13.91 & 42.71 & 54.56 & \cellcolor{maincolor!21} 11.85 & 31.39 & 37.17 & \cellcolor{maincolor!56} 5.78 & 40.0 & 47.5 & \cellcolor{maincolor!23} 7.5 \\
CSOA & 37.51 & 50.06 & \cellcolor{maincolor!17} 12.55 & 41.79 & 53.67 & \cellcolor{maincolor!22} 11.88 & 33.31 & 39.15 & \cellcolor{maincolor!58} 5.84 & \textbf{42.71} & 50.66 & \cellcolor{maincolor!39} 7.95 \\
SACO & 30.37 & 46.69 & \cellcolor{maincolor!78} 16.32 & 38.1 & 52.4 & \cellcolor{maincolor!64} 14.3 & 32.67 & 39.04 & \cellcolor{maincolor!73} 6.37 & 34.96 & 43.78 & \cellcolor{maincolor!71} 8.82 \\
ASCO & 28.9 & 46.12 & \cellcolor{maincolor!93} 17.22 & 37.93 & 51.09 & \cellcolor{maincolor!44} 13.16 & 32.61 & 38.92 & \cellcolor{maincolor!71} 6.31 & 37.4 & 45.78 & \cellcolor{maincolor!55} 8.38 \\
\midrule
Avg. & 34.11 & 48.69 & 14.57 & 41.77 & 54.47 & 12.7 & 33.68 & 39.23 & 5.55 & 39.63 & 47.55 & 7.92 \\
Std. & 3.21 & 1.69 & 2.24 & 1.71 & 1.59 & 1.55 & 1.55 & 1.3 & 0.85 & 2.12 & 2.11 & 0.79 \\

\bottomrule
\end{tabular}%
}
\caption{F1 scores of GPT-4o-mini. The \textbf{best score} for each column is bold.}
\label{tab:acos-asqp-gpt-4o-mini}
\end{table*}

\begin{table*}[t]
\centering
\resizebox{\textwidth}{!}{%
\begin{tabular}{c|ccc|ccc|ccc|ccc}
\toprule
\multirow{2}{*}{\textbf{\textit{Gemini-v1.5}}} & \multicolumn{6}{c|}{\textbf{ASQP}} & \multicolumn{6}{c}{\textbf{ACOS}} \\  
& \multicolumn{3}{c|}{\textbf{Rest15}} & \multicolumn{3}{c|}{\textbf{Rest16}} & \multicolumn{3}{c|}{\textbf{Laptop}} & \multicolumn{3}{c}{\textbf{Rest}} \\ 
\midrule
\textbf{Orders} & \textbf{Origin} & \textbf{Ours} & \textbf{$\Delta$} & \textbf{Origin} & \textbf{Ours} & \textbf{$\Delta$} & \textbf{Origin} & \textbf{Ours} & \textbf{$\Delta$} & \textbf{Origin} & \textbf{Ours} & \textbf{$\Delta$} \\
\midrule
AOSC & 36.56 & 46.73 & \cellcolor{maincolor!30} 10.17 & 44.36 & 56.44 & \cellcolor{maincolor!82} 12.08 & 34.52 & 37.88 & \cellcolor{maincolor!87} 3.36 & 41.0 & 45.47 & \cellcolor{maincolor!32} 4.47 \\
OCSA & 36.74 & 45.98 & \cellcolor{maincolor!10} 9.24 & 46.81 & 58.07 & \cellcolor{maincolor!59} 11.26 & 36.78 & 39.19 & \cellcolor{maincolor!30} 2.41 & 47.51 & 52.38 & \cellcolor{maincolor!56} 4.87 \\
OSAC & 35.16 & 44.66 & \cellcolor{maincolor!16} 9.5 & 44.93 & 56.73 & \cellcolor{maincolor!75} 11.8 & 33.77 & 36.74 & \cellcolor{maincolor!64} 2.97 & 47.29 & 52.87 & \cellcolor{maincolor!100} \textbf{5.58} \\
OSCA & 36.31 & 46.45 & \cellcolor{maincolor!29} 10.14 & 46.5 & 58.3 & \cellcolor{maincolor!75} 11.8 & 36.07 & 38.3 & \cellcolor{maincolor!19} 2.23 & 46.3 & 51.05 & \cellcolor{maincolor!49} 4.75 \\
OACS & 36.28 & 45.58 & \cellcolor{maincolor!11} 9.3 & 45.81 & 57.09 & \cellcolor{maincolor!60} 11.28 & 34.46 & 36.97 & \cellcolor{maincolor!36} 2.51 & \textbf{48.17} & \textbf{53.15} & \cellcolor{maincolor!63} 4.98 \\
AOCS & 38.9 & 48.62 & \cellcolor{maincolor!20} 9.72 & 45.89 & 57.26 & \cellcolor{maincolor!62} 11.37 & 36.78 & 40.08 & \cellcolor{maincolor!84} 3.3 & 44.65 & 48.74 & \cellcolor{maincolor!9} 4.09 \\
COAS & 35.35 & 44.08 & \cellcolor{maincolor!0} 8.73 & 46.26 & 55.42 & \cellcolor{maincolor!0} 9.16 & 35.17 & 37.48 & \cellcolor{maincolor!24} 2.31 & 36.78 & 42.0 & \cellcolor{maincolor!78} 5.22 \\
SAOC & 37.83 & 48.67 & \cellcolor{maincolor!44} 10.84 & 45.94 & 57.65 & \cellcolor{maincolor!72} 11.71 & 33.67 & 35.94 & \cellcolor{maincolor!22} 2.27 & 46.1 & 50.59 & \cellcolor{maincolor!33} 4.49 \\
OASC & 36.14 & 46.01 & \cellcolor{maincolor!23} 9.87 & 45.71 & 57.8 & \cellcolor{maincolor!83} 12.09 & 32.2 & 35.05 & \cellcolor{maincolor!57} 2.85 & 47.23 & 52.72 & \cellcolor{maincolor!94} 5.49 \\
SOAC & 36.42 & 46.53 & \cellcolor{maincolor!28} 10.11 & 45.42 & 56.18 & \cellcolor{maincolor!45} 10.76 & 32.27 & 35.42 & \cellcolor{maincolor!75} 3.15 & 47.63 & 53.06 & \cellcolor{maincolor!90} 5.43 \\
SOCA & 36.47 & 46.3 & \cellcolor{maincolor!22} 9.83 & 46.21 & 56.9 & \cellcolor{maincolor!43} 10.69 & 32.46 & 34.36 & \cellcolor{maincolor!0} 1.9 & 43.95 & 48.9 & \cellcolor{maincolor!61} 4.95 \\
ASOC & 38.14 & 48.35 & \cellcolor{maincolor!30} 10.21 & 45.22 & 57.86 & \cellcolor{maincolor!98} 12.64 & 33.4 & 35.57 & \cellcolor{maincolor!16} 2.17 & 46.3 & 50.24 & \cellcolor{maincolor!0} 3.94 \\
CAOS & 38.87 & 49.83 & \cellcolor{maincolor!46} 10.96 & 46.9 & 59.06 & \cellcolor{maincolor!85} 12.16 & 35.24 & 38.52 & \cellcolor{maincolor!83} 3.28 & 39.6 & 43.53 & \cellcolor{maincolor!0} 3.93 \\
SCAO & 35.9 & 49.42 & \cellcolor{maincolor!100} \textbf{13.52} & 45.3 & 57.91 & \cellcolor{maincolor!98} 12.61 & 33.15 & 36.08 & \cellcolor{maincolor!62} 2.93 & 38.0 & 42.08 & \cellcolor{maincolor!9} 4.08 \\
OCAS & \textbf{39.13} & 48.54 & \cellcolor{maincolor!14} 9.41 & 46.42 & 58.76 & \cellcolor{maincolor!90} 12.34 & \textbf{37.64} & \textbf{40.28} & \cellcolor{maincolor!44} 2.64 & 47.3 & 52.01 & \cellcolor{maincolor!47} 4.71 \\
COSA & 36.4 & 46.29 & \cellcolor{maincolor!24} 9.89 & 44.69 & 53.88 & \cellcolor{maincolor!0} 9.19 & 31.57 & 34.11 & \cellcolor{maincolor!38} 2.54 & 38.26 & 42.56 & \cellcolor{maincolor!22} 4.3 \\
CASO & 38.04 & \textbf{50.14} & \cellcolor{maincolor!70} 12.1 & 46.76 & \textbf{59.14} & \cellcolor{maincolor!91} 12.38 & 33.15 & 36.69 & \cellcolor{maincolor!98} 3.54 & 39.57 & 43.74 & \cellcolor{maincolor!14} 4.17 \\
CSAO & 36.3 & 48.63 & \cellcolor{maincolor!75} 12.33 & 45.38 & 58.06 & \cellcolor{maincolor!100} \textbf{12.68} & 32.41 & 35.97 & \cellcolor{maincolor!100} \textbf{3.56} & 36.74 & 40.92 & \cellcolor{maincolor!15} 4.18 \\
ACOS & 34.65 & 44.54 & \cellcolor{maincolor!24} 9.89 & 44.46 & 56.38 & \cellcolor{maincolor!78} 11.92 & 35.38 & 38.3 & \cellcolor{maincolor!61} 2.92 & 40.63 & 45.04 & \cellcolor{maincolor!29} 4.41 \\
ACSO & 34.92 & 46.53 & \cellcolor{maincolor!60} 11.61 & 43.65 & 54.48 & \cellcolor{maincolor!47} 10.83 & 32.1 & 35.09 & \cellcolor{maincolor!65} 2.99 & 43.55 & 48.22 & \cellcolor{maincolor!44} 4.67 \\
SCOA & 37.56 & 48.04 & \cellcolor{maincolor!36} 10.48 & 45.8 & 55.3 & \cellcolor{maincolor!9} 9.5 & 31.34 & 33.43 & \cellcolor{maincolor!11} 2.09 & 36.56 & 40.75 & \cellcolor{maincolor!15} 4.19 \\
CSOA & 37.93 & 48.53 & \cellcolor{maincolor!39} 10.6 & \textbf{47.42} & 57.04 & \cellcolor{maincolor!13} 9.62 & 30.36 & 32.42 & \cellcolor{maincolor!9} 2.06 & 38.3 & 42.83 & \cellcolor{maincolor!36} 4.53 \\
SACO & 34.51 & 47.77 & \cellcolor{maincolor!94} 13.26 & 41.9 & 53.79 & \cellcolor{maincolor!77} 11.89 & 33.67 & 36.77 & \cellcolor{maincolor!72} 3.1 & 42.08 & 46.97 & \cellcolor{maincolor!58} 4.89 \\
ASCO & 34.19 & 47.23 & \cellcolor{maincolor!89} 13.04 & 43.37 & 54.59 & \cellcolor{maincolor!58} 11.22 & 33.42 & 36.53 & \cellcolor{maincolor!72} 3.11 & 42.0 & 46.68 & \cellcolor{maincolor!45} 4.68 \\
\midrule
\textbf{Avg.} & 36.61 & 47.23 & 10.61 & 45.46 & 56.84 & 11.37 & 33.79 & 36.55 & 2.76 & 42.73 & 47.35 & 4.63 \\
\textbf{Std.} & 1.43 & 1.68 & 1.34 & 1.27 & 1.57 & 1.08 & 1.87 & 1.99 & 0.5 & 4.07 & 4.33 & 0.48 \\
\bottomrule
\end{tabular}%
}
\caption{F1 scores of Gemini-v1.5-Pro. The \textbf{best score} for each column is bold.}
\label{tab:acos-asqp-gemini}
\end{table*}

\begin{table*}[t]
\centering
\resizebox{\textwidth}{!}{%
\begin{tabular}{c|ccc|ccc|ccc|ccc}
\toprule
\multirow{2}{*}{\textbf{\textit{Llama-3.1-70B}}} & \multicolumn{6}{c|}{\textbf{ASQP}} & \multicolumn{6}{c}{\textbf{ACOS}} \\  
& \multicolumn{3}{c|}{\textbf{Rest15}} & \multicolumn{3}{c|}{\textbf{Rest16}} & \multicolumn{3}{c|}{\textbf{Laptop}} & \multicolumn{3}{c}{\textbf{Rest}}  \\ 
\midrule
\textbf{Orders} & \textbf{Origin} & \textbf{Ours} & \textbf{$\Delta$} & \textbf{Origin} & \textbf{Ours} & \textbf{$\Delta$} & \textbf{Origin} & \textbf{Ours} & \textbf{$\Delta$} & \textbf{Origin} & \textbf{Ours} & \textbf{$\Delta$} \\
\midrule
AOSC & 34.97 & 50.5 & \cellcolor{maincolor!35} 15.53 & 42.87 & 57.54 & \cellcolor{maincolor!38} 14.67 & 36.19 & 41.51 & \cellcolor{maincolor!36} 5.32 & 42.54 & 51.28 & \cellcolor{maincolor!3} 8.74 \\
OCSA & 38.35 & 54.35 & \cellcolor{maincolor!43} 16.0 & 46.25 & 60.88 & \cellcolor{maincolor!37} 14.63 & 40.11 & 45.1 & \cellcolor{maincolor!17} 4.99 & 41.82 & 51.78 & \cellcolor{maincolor!41} 9.96 \\
OSAC & 35.53 & 49.33 & \cellcolor{maincolor!7} 13.8 & 45.33 & 58.25 & \cellcolor{maincolor!0} 12.92 & 38.2 & 43.54 & \cellcolor{maincolor!37} 5.34 & 44.11 & 54.1 & \cellcolor{maincolor!41} 9.99 \\
OSCA & \textbf{40.61} & \textbf{57.29} & \cellcolor{maincolor!54} 16.68 & 45.65 & 60.32 & \cellcolor{maincolor!38} 14.67 & 41.02 & \textbf{46.04} & \cellcolor{maincolor!18} 5.02 & 41.79 & 51.31 & \cellcolor{maincolor!27} 9.52 \\
OACS & 38.01 & 51.38 & \cellcolor{maincolor!0} 13.37 & 46.63 & 60.47 & \cellcolor{maincolor!20} 13.84 & 37.79 & 42.65 & \cellcolor{maincolor!9} 4.86 & \textbf{46.13} & \textbf{56.45} & \cellcolor{maincolor!51} 10.32 \\
AOCS & 36.36 & 53.0 & \cellcolor{maincolor!54} 16.64 & 43.83 & 58.56 & \cellcolor{maincolor!39} 14.73 & 34.37 & 40.07 & \cellcolor{maincolor!59} 5.7 & 43.07 & 51.68 & \cellcolor{maincolor!0} 8.61 \\
COAS & 35.82 & 50.5 & \cellcolor{maincolor!21} 14.68 & 46.88 & 61.01 & \cellcolor{maincolor!26} 14.13 & 39.36 & 44.36 & \cellcolor{maincolor!17} 5.0 & 40.6 & 50.52 & \cellcolor{maincolor!39} 9.92 \\
SAOC & 34.53 & 50.62 & \cellcolor{maincolor!44} 16.09 & 41.47 & 57.83 & \cellcolor{maincolor!75} 16.36 & 33.55 & 38.82 & \cellcolor{maincolor!33} 5.27 & 42.48 & 51.77 & \cellcolor{maincolor!20} 9.29 \\
OASC & 36.14 & 50.5 & \cellcolor{maincolor!16} 14.36 & 45.47 & 59.13 & \cellcolor{maincolor!16} 13.66 & 37.75 & 42.63 & \cellcolor{maincolor!10} 4.88 & 45.15 & 55.18 & \cellcolor{maincolor!43} 10.03 \\
SOAC & 34.08 & 48.41 & \cellcolor{maincolor!15} 14.33 & 44.96 & 58.75 & \cellcolor{maincolor!19} 13.79 & 35.92 & 41.35 & \cellcolor{maincolor!43} 5.43 & 45.0 & 54.83 & \cellcolor{maincolor!37} 9.83 \\
SOCA & 37.88 & 55.1 & \cellcolor{maincolor!63} 17.22 & 46.01 & 60.91 & \cellcolor{maincolor!43} 14.9 & 38.69 & 44.35 & \cellcolor{maincolor!56} 5.66 & 41.56 & 51.54 & \cellcolor{maincolor!41} 9.98 \\
ASOC & 33.1 & 49.76 & \cellcolor{maincolor!54} 16.66 & 42.73 & 58.49 & \cellcolor{maincolor!62} 15.76 & 35.7 & 41.8 & \cellcolor{maincolor!82} 6.1 & 42.54 & 52.93 & \cellcolor{maincolor!54} 10.39 \\
CAOS & 38.03 & 54.15 & \cellcolor{maincolor!45} 16.12 & 45.98 & 60.8 & \cellcolor{maincolor!41} 14.82 & 36.45 & 42.51 & \cellcolor{maincolor!80} 6.06 & 37.22 & 46.49 & \cellcolor{maincolor!20} 9.27 \\
SCAO & 33.79 & 51.74 & \cellcolor{maincolor!75} 17.95 & 44.56 & 60.65 & \cellcolor{maincolor!69} 16.09 & 36.58 & 41.98 & \cellcolor{maincolor!41} 5.4 & 36.45 & 46.96 & \cellcolor{maincolor!57} 10.51 \\
OCAS & 38.65 & 53.32 & \cellcolor{maincolor!21} 14.67 & \textbf{47.46} & \textbf{63.6} & \cellcolor{maincolor!70} 16.14 & \textbf{41.28} & 45.98 & \cellcolor{maincolor!0} 4.7 & 41.86 & 51.94 & \cellcolor{maincolor!44} 10.08 \\
COSA & 37.24 & 52.92 & \cellcolor{maincolor!38} 15.68 & 46.24 & 59.76 & \cellcolor{maincolor!13} 13.52 & 39.66 & 45.0 & \cellcolor{maincolor!37} 5.34 & 38.14 & 47.56 & \cellcolor{maincolor!24} 9.42 \\
CASO & 34.22 & 53.64 & \cellcolor{maincolor!100} \textbf{19.42} & 43.61 & 61.08 & \cellcolor{maincolor!100} \textbf{17.47} & 36.08 & 42.47 & \cellcolor{maincolor!100} \textbf{6.39} & 35.66 & 46.08 & \cellcolor{maincolor!55} 10.42 \\
CSAO & 33.23 & 51.04 & \cellcolor{maincolor!73} 17.81 & 42.56 & 59.65 & \cellcolor{maincolor!91} 17.09 & 33.27 & 39.26 & \cellcolor{maincolor!76} 5.99 & 35.16 & 45.89 & \cellcolor{maincolor!64} 10.73 \\
ACOS & 35.09 & 51.22 & \cellcolor{maincolor!45} 16.13 & 43.63 & 58.56 & \cellcolor{maincolor!44} 14.93 & 36.67 & 42.4 & \cellcolor{maincolor!60} 5.73 & 40.2 & 50.49 & \cellcolor{maincolor!51} 10.29 \\
ACSO & 32.52 & 51.39 & \cellcolor{maincolor!90} 18.87 & 41.95 & 58.63 & \cellcolor{maincolor!82} 16.68 & 34.82 & 40.94 & \cellcolor{maincolor!84} 6.12 & 38.97 & 50.87 & \cellcolor{maincolor!100} \textbf{11.9} \\
SCOA & 36.19 & 51.45 & \cellcolor{maincolor!31} 15.26 & 44.59 & 58.05 & \cellcolor{maincolor!11} 13.46 & 39.23 & 44.31 & \cellcolor{maincolor!22} 5.08 & 41.42 & 50.71 & \cellcolor{maincolor!20} 9.29 \\
CSOA & 36.3 & 51.74 & \cellcolor{maincolor!34} 15.44 & 44.67 & 59.2 & \cellcolor{maincolor!35} 14.53 & 38.03 & 43.86 & \cellcolor{maincolor!66} 5.83 & 40.13 & 49.72 & \cellcolor{maincolor!29} 9.59 \\
SACO & 31.68 & 50.0 & \cellcolor{maincolor!81} 18.32 & 42.66 & 58.84 & \cellcolor{maincolor!71} 16.18 & 34.81 & 40.61 & \cellcolor{maincolor!65} 5.8 & 38.83 & 50.17 & \cellcolor{maincolor!82} 11.34 \\
ASCO & 32.84 & 51.04 & \cellcolor{maincolor!79} 18.2 & 44.51 & 59.88 & \cellcolor{maincolor!53} 15.37 & 35.5 & 40.79 & \cellcolor{maincolor!34} 5.29 & 41.51 & 52.68 & \cellcolor{maincolor!77} 11.17 \\
\midrule
Avg. & 35.63 & 51.85 & 16.22 & 44.6 & 59.62 & 15.01 & 37.13 & 42.6 & 5.47 & 40.93 & 50.96 & 10.02 \\
Std. & 2.25 & 2.03 & 1.63 & 1.66 & 1.4 & 1.23 & 2.26 & 2.01 & 0.46 & 2.94 & 2.82 & 0.78 \\
\bottomrule
\end{tabular}%
}
\caption{F1 scores of Llama-3.1-70B-Instruct. The \textbf{best score} for each column is bold.}
\label{tab:acos-asqp-llama3.1}
\end{table*}

\end{document}